\documentclass[final]{cvpr}

\usepackage{times}
\usepackage{epsfig}
\usepackage{graphicx}
\usepackage{amsmath}
\usepackage{amssymb}
\usepackage{enumitem}
\usepackage{subfigure}
\usepackage{multirow}
\setlist{nosep}
\usepackage{amssymb}
\usepackage{pifont}
\newcommand{\cmark}{\ding{51}}%
\usepackage{diagbox}
\usepackage[table,xcdraw,dvipsnames]{xcolor}

\def\ie{\emph{i.e.}}
\def\eg{\emph{e.g.}}

\usepackage[pagebackref=true,breaklinks=true,colorlinks,bookmarks=false]{hyperref}

\def\cvprPaperID{2} 
\def\confYear{CVPR 2022}

\begin{document}

\title{UIGR: Unified Interactive Garment Retrieval}

\author{Xiao Han$^{1,2}$ Sen He$^{1,2}$ Li Zhang$^3$ Yi-Zhe Song$^{1,2}$ Tao Xiang$^{1,2}$ \\
$^1$CVSSP, University of Surrey \\ 
$^2$iFlyTek-Surrey Joint Research Centre on Artificial Intelligence \\
$^3$School of Data Science, Fudan University \\
{\tt\small \{xiao.han, sen.he, y.song, t.xiang\}@surrey.ac.uk, lizhangfd@fudan.edu.cn}
}

\maketitle

\begin{abstract}
Interactive garment retrieval (IGR) aims to retrieve a target garment image based on a reference garment image along with user feedback on what to change on the reference garment. 
Two IGR tasks have been studied extensively: text-guided garment retrieval (TGR) and visually compatible garment retrieval (VCR). 
The user feedback for the former indicates what semantic attributes to change with the garment category preserved, while the category is the only thing to be changed explicitly for the latter, with an implicit requirement on style preservation. 
Despite the similarity between these two tasks and the practical need for an efficient system tackling both, they have never been unified and modeled jointly.
In this paper, we propose a \textbf{U}nified \textbf{I}nteractive \textbf{G}arment \textbf{R}etrieval (UIGR) framework to unify TGR and VCR.
To this end, we first contribute a large-scale benchmark suited for both problems. 
We further propose a strong baseline architecture to integrate TGR and VCR in one model.
Extensive experiments suggest that unifying two tasks in one framework is not only more efficient by requiring a single model only, it also leads to better performance.
Code and datasets are available at \href{https://github.com/BrandonHanx/CompFashion}{GitHub}.
\vspace*{-4mm}
\end{abstract}
\section{Introduction}

In computer vision, there is a long line of research on understanding garment image content \cite{liu2016deepfashion,wu2021fashioniq,vasileva2018learning,han2018viton,al2017fashionforecasting,cheng2021survey}.
Among them, \textbf{I}nteractive \textbf{G}arment \textbf{R}etrieval (IGR)  \cite{zhao2017memory,vo2019tirg,wu2021fashioniq} is most relevant to the garment search problem. IGR aims to retrieve a target garment image based on a reference garment image along with user feedback on what to change on the reference garment. It enables a shopper to find exactly what she/he wants because it allows for the fine-tuning of search results through user feedback.  

\begin{figure}[t]
\begin{center}
\includegraphics[width=0.9\linewidth]{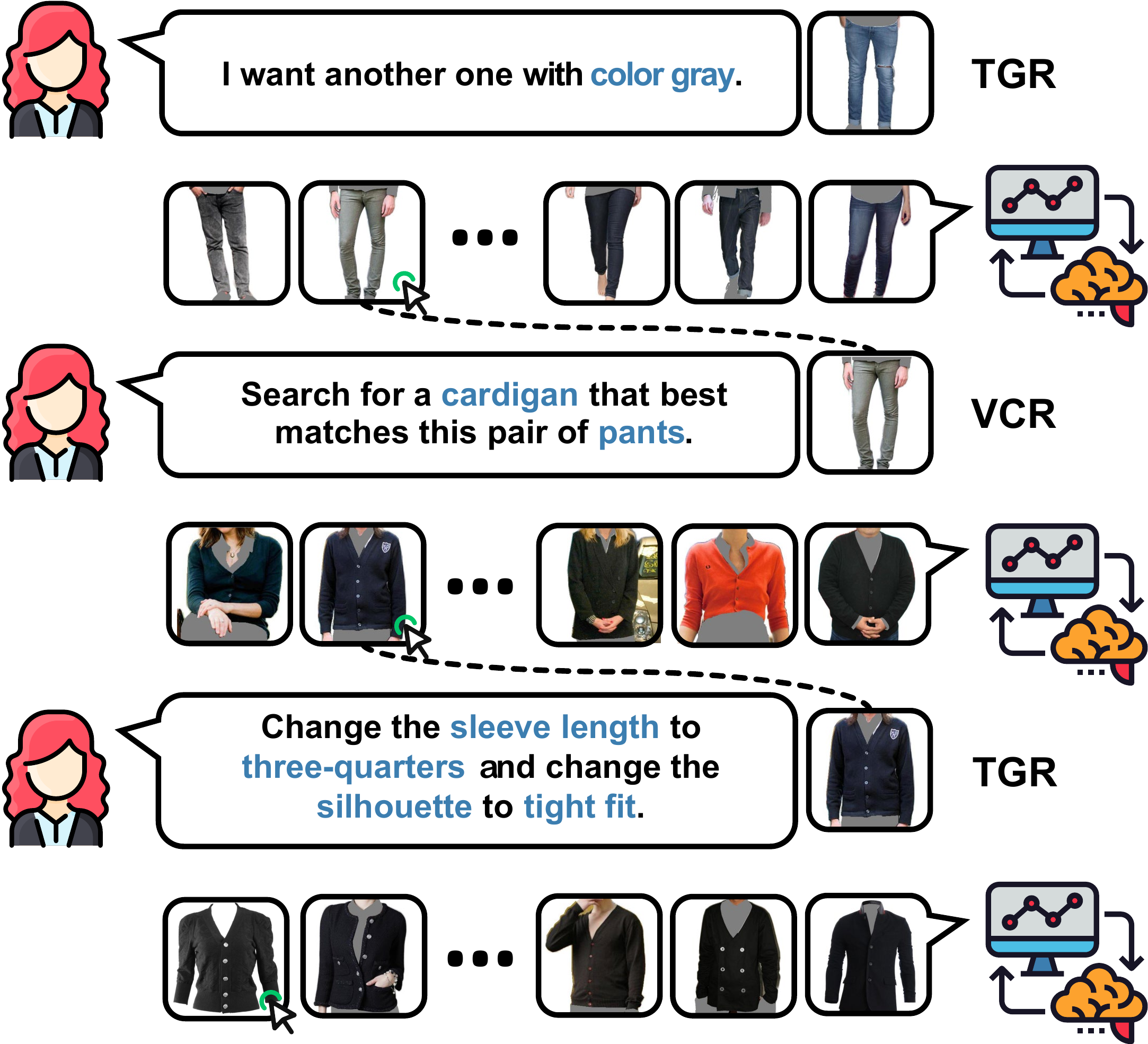}
\end{center}
\vspace*{-2mm}
\caption{An example IGR scenario where both TGR and VCR can take place.
Given a reference garment, users may search for a garment of the same category with some attribute changes (TGR), or  visually compatible garments of different categories (VCR).
}
\label{fig:introduction}
\vspace*{-2mm}
\end{figure}

Two IGR tasks, namely \textbf{T}ext-\textbf{G}uided garment \textbf{R}etrieval (TGR) \cite{vo2019tirg,liu2021cirr} and \textbf{V}isually \textbf{C}ompatible garment \textbf{R}etrieval (VCR) \cite{lin2020fashion,hou2021disentangled} have been studied so far (see Figure \ref{fig:introduction}).
TGR (dialog 1 \& 3 of Figure \ref{fig:introduction}) retrieves garments in the same category as the reference garment. 
The feedback is in the form of either synthetic sentence \cite{guo2018dialog, vo2019tirg} or natural language \cite{wu2021fashioniq}, indicating the intended attribute changes from the reference to the target garment. 
In contrast, the feedback for VCR (dialog 2 of Figure \ref{fig:introduction}) typically only indicates category change explicitly, in the form of an indicator rather than text \cite{lin2020fashion,hou2021disentangled}. 
Nevertheless, as the retrieval is constrained to only visually compatible items, implicit feedback is to preserve the style so that the reference and target look lovely when worn together.

Despite being two instantiations of IGR,  TGR and VCR have never been studied together in a unified framework.
Indeed, they are evaluated on completely different sets of benchmarks. 
The developed methods also seem pretty different. 
TGR is usually done by first compositing the reference garment with the interaction signal together and then retrieving the garments similar to the composited query \cite{vo2019tirg,chen2020val,lee2021cosmo,shin2021rtic}. 
Since different garments are compatible along multiple dimensions, such as color, pattern, and material, previous works in VCR typically learn subspace embeddings to capture different notions of similarity and aim to learn a joint embedding space where compatible garments of different categories are close \cite{lin2020fashion,hou2021disentangled}.

In this paper, for the first time, we propose \textbf{U}nified \textbf{I}nteractive \textbf{G}arment \textbf{R}etrieval (UIGR) to unify the two tasks in a single framework. 
We argue that there are two benefits for doing so:
(1) As shown in Figure \ref{fig:introduction}, it is common to have both tasks incurred in the same shopping session. It is thus more efficient to build one rather than two separate models to tackle both tasks. 
(2) Due to the similarity in format (\ie, both are IGR tasks), having a single multi-task framework makes it possible for both tasks to benefit from each other when trained jointly end-to-end. 
However, unifying the two tasks is challenging, with two main obstacles to overcome: the lack of benchmarks and the discrepancy in the two types of user feedback.

To this end, we try to solve these problems with two main contributions:
(1) We establish a novel benchmark for the study of this unified problem by re-purposing Fashionpedia \cite{jia2020fashionpedia},
where prompt engineering is adopted to generate user feedback from fine-grained attributes.
(2) We introduce a multi-task model jointly learning two tasks, which unifies TGR and VCR in a single framework and serves as a strong baseline for UIGR. 
Experiments demonstrate that unifying the two tasks in a single model is not only possible but also yields better overall performance, compared with modeling them separately using two models.

\section{Related work}
\noindent \textbf{Text-guided garment retrieval.}
TGR is a special type of image retrieval problem with multimodal compositional queries \cite{vo2019tirg,liu2021cirr,changpinyo2021mousetrace}. 
In general, the user feedback used to guide the searching process can be attributes \cite{zhao2017memory,han2017automatic,ak2018learning}, synthetic sentences \cite{guo2018dialog, vo2019tirg}, and natural language (free text) \cite{wu2021fashioniq, yuan2021turn}.
Different TGR models proposed so far differ primarily in the design of their compositors.
A compositor plays a fundamental role to integrate the textual information with the imagery modality.
TGR compositors have been proposed based on various techniques, such as gating mechanism \cite{vo2019tirg}, hierarchical attention \cite{chen2020val,jandial2020trace,dodds2020maaf,hosseinzadeh2020composed}, graph neural network \cite{zhang2020joint,shin2021rtic}, joint learning \cite{chen2020jvsm,kim2021dcnet,shin2021rtic,yang2021cross,zhang2021heterogeneous}, ensemble learning \cite{wen2021clvc}, style-content modification \cite{lee2021cosmo,chawla2021leveraging} and vision \& language pre-training \cite{liu2021cirr}.
\begin{figure*}[ht]
\begin{center}
\includegraphics[width=0.9\linewidth]{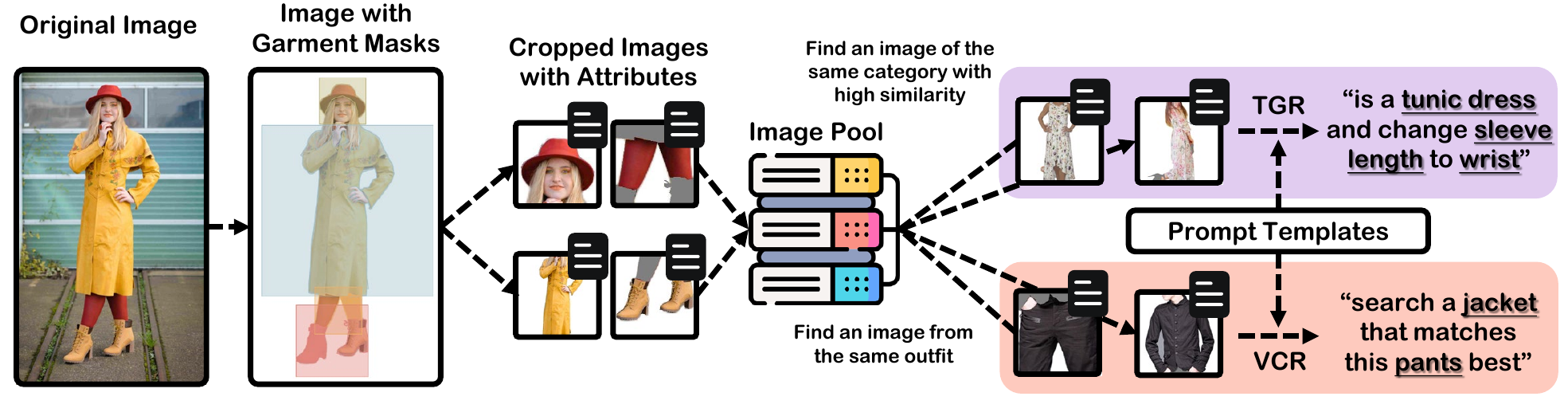}
\end{center}
\vspace{-2mm}
\caption{Overview of the dataset collection process. 
The whole pipeline is based on the image and corresponding high-quality annotations from Fashionpedia \cite{jia2020fashionpedia}. 
(1) We firstly construct an image pool by cropping each garment using its ground truth mask. 
(2) To construct TGR triplets, we select a pair of images with the same category and high similarity. 
Then the user feedback is generated by filling relative attributes in the blank of prompt templates. 
(3) For VCR triplets construction, the image pair is selected according to whether both images are from the same outfit.
We generate this kind of user feedback by mentioning the categories of both reference and target images.}
\label{fig:data_pipeline}
\vspace{-2mm}
\end{figure*}

\vspace{1mm}
\noindent \textbf{Visually compatible garment retrieval.} 
Predicting fashion compatibility is to determine whether two garments of different categories match well aesthetically.
On this basis, the recommendation can be done either as fill-in-the-blank \cite{han2017learning} at item level or as personalized outfit recommendation \cite{lu2019hash,lu2021personalized} at outfit level.
In addition to being a set, an outfit can also be represented as a sequence \cite{han2017learning}, or a graph \cite{cucurull2019context}.

Instead of computing the compatibility in a single space, most approaches \cite{veit2017csn,vasileva2018learning,tan2019learning,lin2020fashion,hou2021disentangled,kim2021self} explore learning subspace embeddings to capture different notions of compatibility. 
\cite{veit2017csn,vasileva2018learning} learn many conditional subspaces, each for a pair of categories. 
\cite{tan2019learning} learns several subspaces conditioned on the features from both the reference garment and the target garment. 
However, this kind of method is not suitable for large-scale retrieval where exhaustive comparison is prohibitive.
\cite{lin2020fashion,hou2021disentangled} concatenate one-hot labels of the reference and target category to represent the interaction signal to meet the setting of large-scale retrieval.


\vspace{1mm}
\noindent \textbf{Fashion datasets.}
Over the past few years, many fashion datasets have been proposed for multiple applications \cite{cheng2021survey}, such as detection \cite{liu2016deepfashion,liu2016landmark,ge2019deepfashion2}, retrieval \cite{liu2016deepfashion,ge2019deepfashion2,ma2020fine}, attribute recognition \cite{liu2016deepfashion,han2017automatic}, popularity learning \cite{lo2019dressing,ma2019who,al2017fashionforecasting} and synthesis \cite{han2018viton,jiang2020psgan}. 
The most related datasets to our work are \cite{guo2018dialog,han2017automatic,wu2021fashioniq} for TGR and \cite{vasileva2018learning,lin2020fashion,revanur2021semi} for VCR.
Besides not being suitable for the unified setting, previous TGR and VCR benchmarks have some other problems, which will be explained in next Section.

\section{New benchmark for IGR}
\label{par:data_preparation}
Next, we describe the data collection process and provide an in-depth analysis of UIGR. 
The overall data collection procedure is illustrated in Figure \ref{fig:data_pipeline}. 
The basic statistics is summarized in Table \ref{tab:dataset_statistics} and \ref{tab:dataset_comparsions} \footnote{``Triplet" in this article refers to one piece of data, \ie, two images and one sentence, rather than anchor, positive and negative sample pair.}.

\subsection{Image and attribute collection}
We collect UIGR garment images based on the original images, garment bounding boxes, garment segmentation masks, and fine-grained attributes from Fashionpedia \cite{jia2020fashionpedia} with a series of pre-processing
\footnote{More details about pre-processing steps are listed in Supp. Mat.}.


\begin{figure}[t]
\centering
\includegraphics[width=0.8\linewidth]{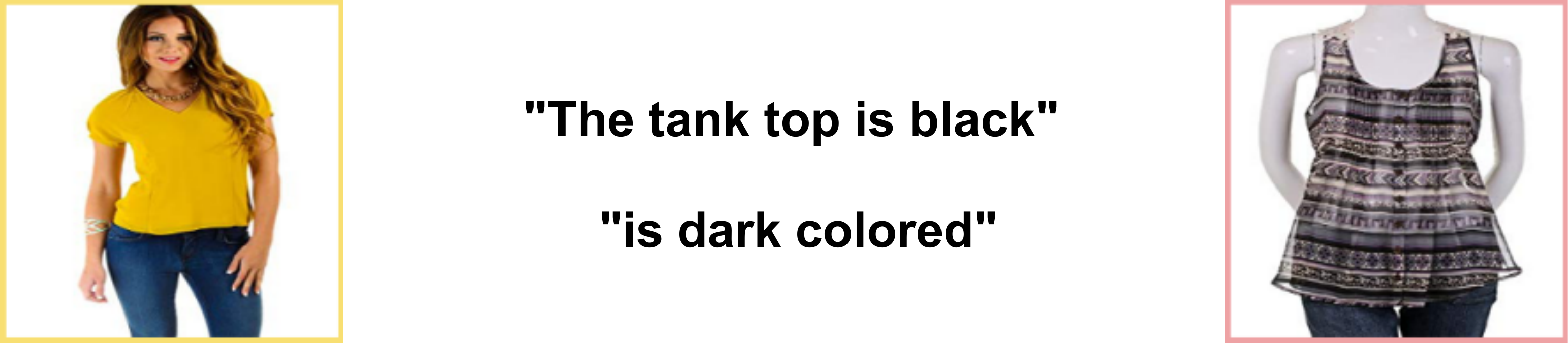}
\includegraphics[width=0.8\linewidth]{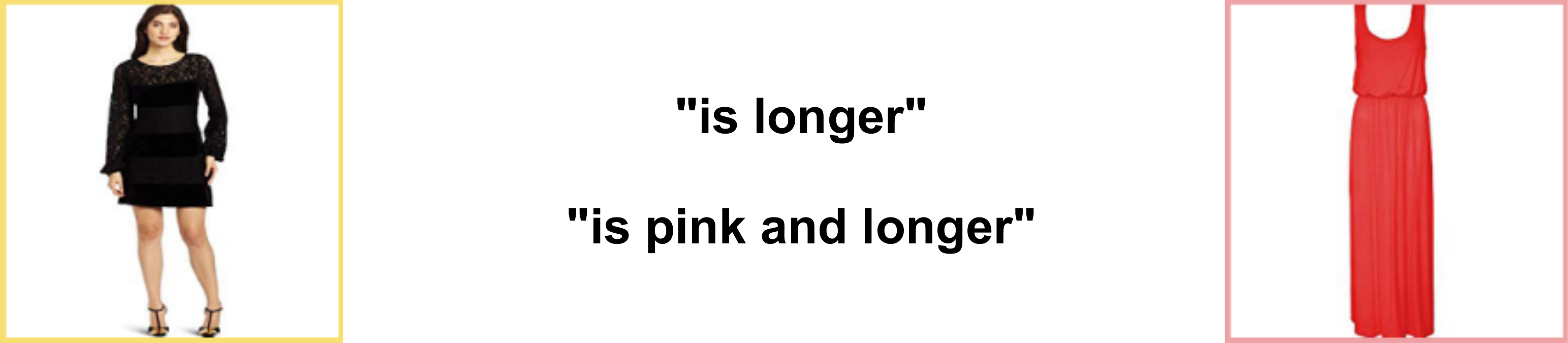}
\caption{Typical bad triplets in FashionIQ \cite{wu2021fashioniq}.}
\label{fig:fashioniq_examples}
\end{figure}
\begin{figure}[t]
\centering
\includegraphics[width=0.8\linewidth]{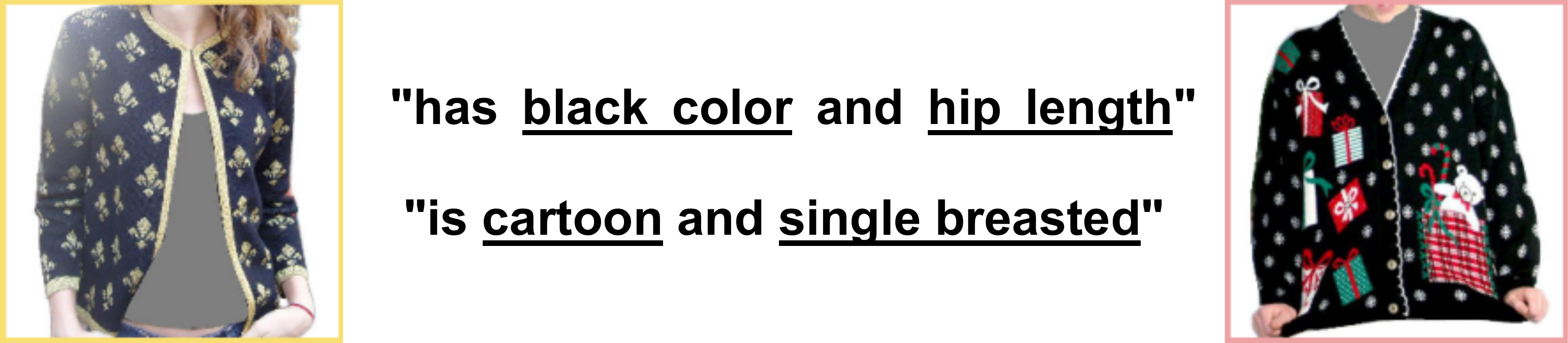}
\includegraphics[width=0.8\linewidth]{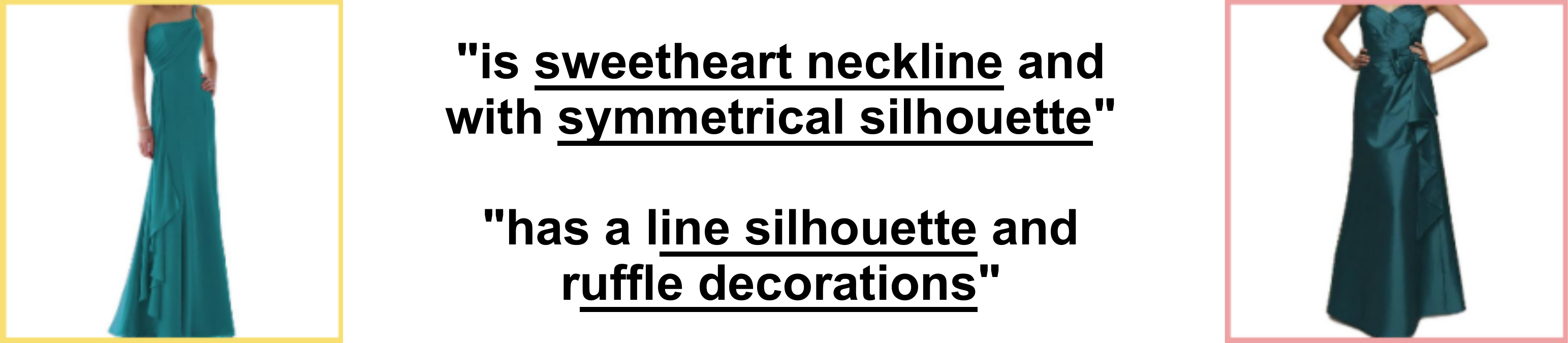}
\caption{Triplet examples in UIGR TGR subset.}
\label{fig:uigr_tgr_examples}
\end{figure}

\begin{figure}[t]
\centering
\includegraphics[width=0.8\linewidth]{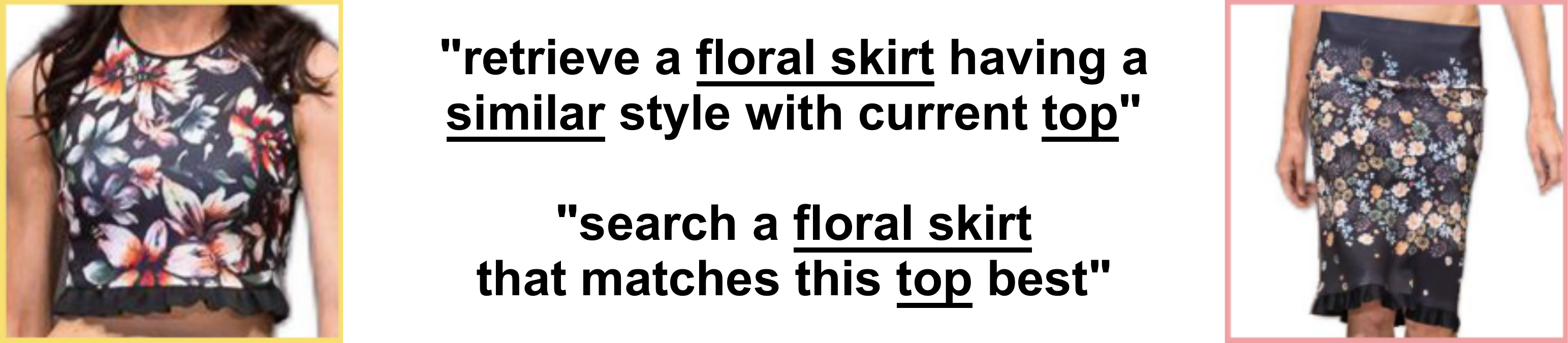}
\includegraphics[width=0.8\linewidth]{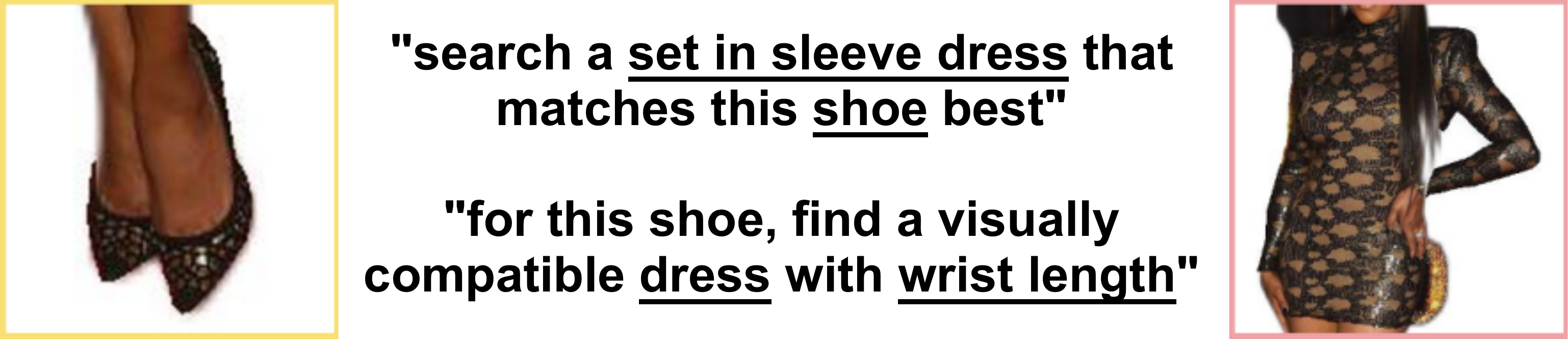}
\caption{Triplet examples in UIGR VCR subset.}
\label{fig:uigr_vcr_examples}
\end{figure}

\subsection{Image pair selection}
\noindent \textbf{TGR subset.} 
Previous benchmarks \cite{han2017automatic,wu2021fashioniq} select image pairs by comparing the similarity of text information, \eg,  image titles or attribute labels.
As shown in Figure \ref{fig:fashioniq_examples},
this selection strategy often leads to weakly related image pairs with drastically different visual appearances.
The user feedback thus cannot accurately describe all the changes necessary to align the image pairs because there are too many changes needed.
We thus take a different strategy: using image similarity instead of text similarity for pair selection.
Specifically, we use a DenseNet \cite{huang2017densenet,kim2021dcnet} pre-trained on DeepFashion \cite{liu2016deepfashion} to get image feature vectors.
Next, for each image, we calculate the cosine similarity between it and all images of the same category in the image pool and only consider the top three most similar matches.

\vspace{1mm}
\noindent \textbf{VCR subset.} 
We select all garments coming from the same outfit in a bidirectional way to construct image pairs, which is a standard procedure adopted in previous VCR benchmarks \cite{vasileva2018learning,lin2020fashion}.

\begin{table}[t]
\begin{center}
\resizebox{0.8\columnwidth}{!}{
\begin{tabular}{ccccc}
\hline
\multirow{2}{*}{\textbf{Split}} & \multirow{2}{*}{\textbf{\# Images}} & \multirow{2}{*}{\textbf{\# Outfits}} & \multicolumn{2}{c}{\textbf{\# Triplets}}                        \\ \cline{4-5} 
                                &                                     &                                      & \textbf{TGR} & \multicolumn{1}{c}{\textbf{VCR}} \\  \hline 
Train           & 76,685         & 29,321            & 210,189      & 190,150                    \\ 
Validation      & 25,181         & 9,688             & 68,847       & 61,776                    \\ 
Test            & 25,434         & 9,814             & 69,639       & 62,418                  \\ \hline
\end{tabular}
}
\end{center}
\vspace*{-2mm}
\caption{Dataset statistics of UIGR.}
\label{tab:dataset_statistics}
\vspace*{-2mm}
\end{table}
\begin{table}[t]
\begin{center}
\resizebox{\columnwidth}{!}{
\begin{tabular}{cccc}
\hline
\textbf{Dataset}           & \textbf{\# Triplets} & \textbf{\# Categories} & \textbf{Caption length} \\ \hline 
Shoes \cite{berg2010shoes,guo2018dialog}                  & 10k                     & 1                         & 5.22 words                    \\ 
Fashion200K \cite{han2017automatic,vo2019tirg}           & 172k                    & 5                         & 4.00 words                    \\ 
FashionIQ \cite{wu2021fashioniq}             & 18k                     & 3                         & 5.36 words                   \\ 
Our TGR       & 381k                    & 27                        & 6.33 words                   \\ \hline 
\textbf{Dataset}  & \textbf{\# Outfits}  & \textbf{\# Categories} & \textbf{Interaction signal}       \\ \hline  
Polyvore retrieval \cite{lin2020fashion}     & 17k                     & 16                        & One-hot labels                 \\ 
Our VCR       & 49k                     & 27                        & Text                          \\ \hline
\end{tabular}
}
\end{center}
\vspace*{-2mm}
\caption{Comparisons with other related datasets.}
\label{tab:dataset_comparsions}
\vspace*{-4mm}
\end{table}

\subsection{User feedback generation}
Because the scale of UIGR is more than twenty times that of FashionIQ, manually annotating each image pair with fine-grained user feedback is laborious and costly.
To this end, we adopt prompt engineering to automatically generate the user feedback based on the relative attributes between two garments.
Following the setting of FashionIQ, we generate two sentences for each image pair.

\vspace{1mm}
\noindent \textbf{TGR subset.} 
We manually summarize tens of cloze prompt templates from FashionIQ captions.
These templates include several single phrases, such as \textit{``has \{V\} \{A\}"} and \textit{``change \{A\} to \{V\}"}, where \{V\} and \{A\} hold the blank for one attribute name and its value.
The templates of multiple phrases are based on the combination of single phrases.
Finally, the relative attributes between two images are filled in the blanks of the randomly selected prompt template.

\vspace{1mm}
\noindent \textbf{VCR subset.} 
To unify the VCR task with TGR, they need to have the same user feedback format, \ie, sentences describing the intended changes to the reference garment. 
One obvious choice is to use the prompt engineering technique to generate sentences describing only the category changes for VCR. However, this fails to capture the implicit user feedback when it comes to VCR. That is, the style of the target garment needs to be consistent with that of the reference. 

To this end,  we first calculate the correlation matrix of all attributes between any two kinds of garments. 
When constructing VCR triplets, we will predict the most likely target attributes based on the existing attributes of the reference image.
Next, we will randomly mention one attribute in the predicted attributes using the attribute correlation matrix when generating user feedback.

Different from the TGR subset, we manually design several prompt templates for VCR, such as
\textit{``search a \{TV\} \{TC\} that matches this \{RC\} best"}
and
\textit{``for this \{RC\}, find a visually compatible \{TV\} \{TC\}"},
where \{TV\}, \{TC\} and \{RC\} stand for the target attribute value, target category and reference category, respectively.

\subsection{Dataset analysis}
\label{par:dataset_statistics}


The examples of our collected TGR triplets are depicted in Figure \ref{fig:uigr_tgr_examples}.
Compared with those from FashionIQ in Figure \ref{fig:fashioniq_examples}, our triplets seems  more reasonable.
In particular, although all relative captions in FashionIQ are annotated via a crowdsourcing platform, many captions are too ambiguous to describe the exact search direction. 
Since we select image pairs based on the image similarity to avoid significant visual changes, the subsequently generated user feedback is more accurate and fine-grained.

Figure \ref{fig:uigr_vcr_examples} shows the examples of VCR subset
Compared with one-hot labels for user feedback, sentences are more flexible and scalable to integrate more fine-grained information from users.
Further, the VCR task now has the same setting as the TGR, making unification possible.

\section{Experiments}

\begin{figure}[t]
\begin{center}
\includegraphics[width=\linewidth]{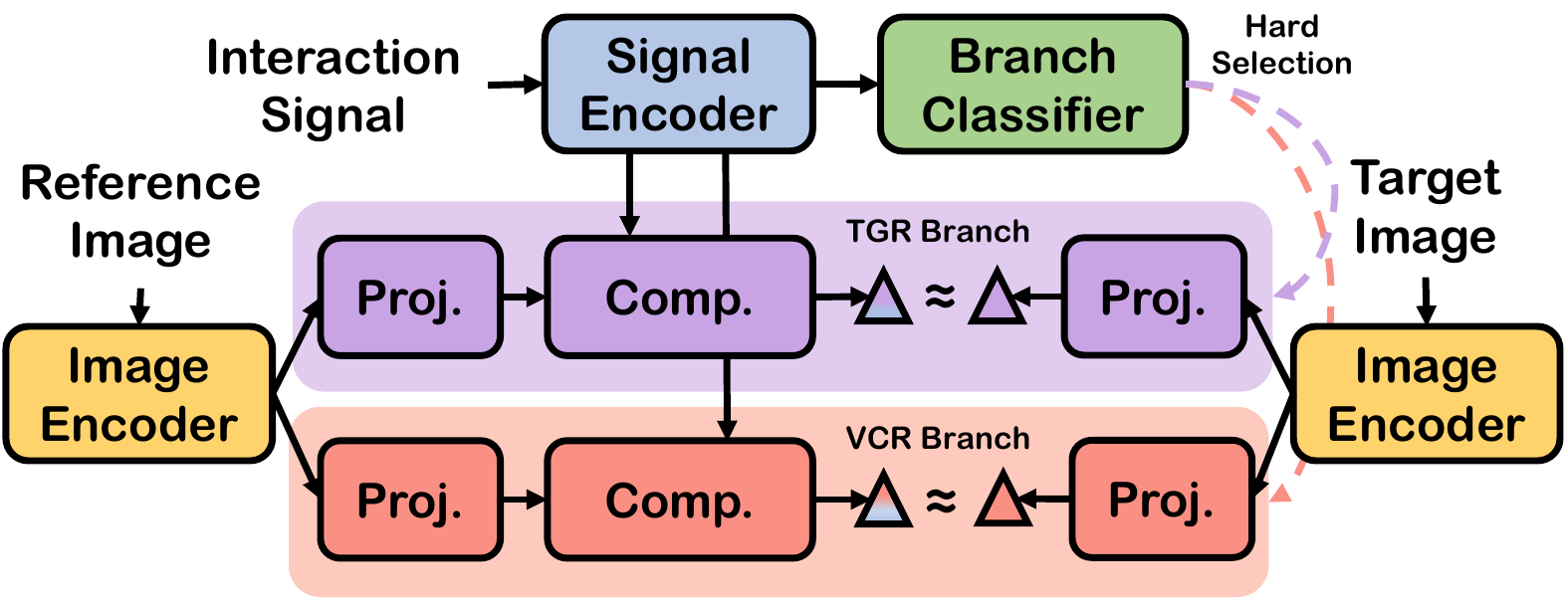}
\end{center}
\vspace*{-3mm}
\caption{Proposed multi-task architecture for UIGR.}
\label{fig:proposed_pipeline}
\vspace*{-4mm}
\end{figure}

Although there are different implementations for the compositors of VCR and TGR, they share the same goal: preserving unmentioned visual appearance aspects of the reference and changing only those mentioned in the interaction signal/feedback.
Our multi-task model unifies the two tasks based on the same goal. 
However, to accommodate the major difference in the change directions of the two tasks, namely whether the category is preserved or changed, we use different compositors. 
As shown in Figure \ref{fig:proposed_pipeline}, two branches are used for separately learning two composition processes with shared image and signal encoders.
Considering that the features needed to be modified for the two branches are not the same, we use two projection modules to project image features to two latent spaces ahead of the composition process. 
We also jointly learn a classifier to distinguish different user feedback.
With it, our model can automatically determine which branch should be selected to do composition during inference, thus allowing the real-world application scenario depicted in Figure \ref{fig:introduction} to be supported by one model.


\label{par:experiments}

\begin{table}[t]
\begin{center}
\resizebox{\linewidth}{!}{
\begin{tabular}{|cc||cccccccc|}
\hline
\multicolumn{2}{|c||}{}                                                                      & \multicolumn{3}{c|}{\textbf{TGR Results}}                                                                          & \multicolumn{3}{c|}{\textbf{VCR Results}}                                                                          & \multicolumn{2}{c|}{\textbf{Mean}}                                              \\ \cline{3-10}
\multicolumn{2}{|c||}{\multirow{-2}{*}{\diagbox{\textbf{Comp.}}{\textbf{Metrics}}}}                                 & \textbf{R@10}                 & \textbf{R@50}                 & \multicolumn{1}{c|}{\textbf{mAP}}                  & \textbf{R@10}                 & \textbf{R@50}                 & \multicolumn{1}{c|}{\textbf{mAP}}                  & \textbf{R@K}                           & \textbf{mAP}                                             \\ \hline \hline
\multicolumn{1}{|c|}{}                                 & \textbf{I}                         & 38.98                         & 72.08                         & \multicolumn{1}{c|}{14.29}                         & 71.03                         & 86.83                         & \multicolumn{1}{c|}{46.82}                         & \textbf{67.23}                         & \textbf{30.56}                          \\
\multicolumn{1}{|c|}{\multirow{-2}{*}{\textbf{CSA} \cite{lin2020fashion}}}   & \cellcolor[HTML]{ECF4FF}\textbf{U} & \cellcolor[HTML]{ECF4FF}36.90 & \cellcolor[HTML]{ECF4FF}70.57 & \multicolumn{1}{c|}{\cellcolor[HTML]{ECF4FF}13.37} & \cellcolor[HTML]{ECF4FF}70.46 & \cellcolor[HTML]{ECF4FF}86.88 & \multicolumn{1}{c|}{\cellcolor[HTML]{ECF4FF}46.47} & \cellcolor[HTML]{ECF4FF}66.20          & \cellcolor[HTML]{ECF4FF}29.92          \\ \hline
\multicolumn{1}{|c|}{}                                 & \textbf{I}                         & 46.27                         & 77.57                         & \multicolumn{1}{c|}{19.78}                         & 69.30                         & 85.88                         & \multicolumn{1}{c|}{46.15}                         & 69.76                                  & 32.97                                  \\
\multicolumn{1}{|c|}{\multirow{-2}{*}{\textbf{TIRG} \cite{vo2019tirg}}}  & \cellcolor[HTML]{ECF4FF}\textbf{U} & \cellcolor[HTML]{ECF4FF}45.06 & \cellcolor[HTML]{ECF4FF}76.75 & \multicolumn{1}{c|}{\cellcolor[HTML]{ECF4FF}18.91} & \cellcolor[HTML]{ECF4FF}72.11 & \cellcolor[HTML]{ECF4FF}88.42 & \multicolumn{1}{c|}{\cellcolor[HTML]{ECF4FF}48.54} & \cellcolor[HTML]{ECF4FF}\textbf{70.59} & \cellcolor[HTML]{ECF4FF}\textbf{33.73} \\ \hline
\multicolumn{1}{|c|}{}                                 & \textbf{I}                         & 43.27                         & 75.10                         & \multicolumn{1}{c|}{18.06}                         & 62.99                         & 81.97                         & \multicolumn{1}{c|}{40.40}                         & 65.83                                  & 29.23                                  \\
\multicolumn{1}{|c|}{\multirow{-2}{*}{\textbf{VAL} \cite{chen2020val}}}   & \cellcolor[HTML]{ECF4FF}\textbf{U} & \cellcolor[HTML]{ECF4FF}40.19 & \cellcolor[HTML]{ECF4FF}71.78 & \multicolumn{1}{c|}{\cellcolor[HTML]{ECF4FF}17.28} & \cellcolor[HTML]{ECF4FF}67.95 & \cellcolor[HTML]{ECF4FF}86.24 & \multicolumn{1}{c|}{\cellcolor[HTML]{ECF4FF}44.72} & \cellcolor[HTML]{ECF4FF}\textbf{66.54} & \cellcolor[HTML]{ECF4FF}\textbf{31.00} \\ \hline
\multicolumn{1}{|c|}{}                                 & \textbf{I}                         & 40.24                         & 72.15                         & \multicolumn{1}{c|}{17.31}                         & 64.10                         & 83.18                         & \multicolumn{1}{c|}{41.36}                         & 64.92                                  & 29.34                                   \\
\multicolumn{1}{|c|}{\multirow{-2}{*}{\textbf{CoSMo} \cite{lee2021cosmo}}} & \cellcolor[HTML]{ECF4FF}\textbf{U} & \cellcolor[HTML]{ECF4FF}40.96 & \cellcolor[HTML]{ECF4FF}72.40 & \multicolumn{1}{c|}{\cellcolor[HTML]{ECF4FF}17.62} & \cellcolor[HTML]{ECF4FF}68.64 & \cellcolor[HTML]{ECF4FF}86.41 & \multicolumn{1}{c|}{\cellcolor[HTML]{ECF4FF}45.05} & \cellcolor[HTML]{ECF4FF}\textbf{67.10} & \cellcolor[HTML]{ECF4FF}\textbf{31.36} \\ \hline
\multicolumn{1}{|c|}{}                                 & \textbf{I}                         & 48.23                         & 78.79                         & \multicolumn{1}{c|}{19.98}                         & 69.28                         & 86.26                         & \multicolumn{1}{c|}{46.10}                         & 70.64                                  & 33.04                                   \\
\multicolumn{1}{|c|}{\multirow{-2}{*}{\textbf{RTIC} \cite{shin2021rtic}}}  & \cellcolor[HTML]{ECF4FF}\textbf{U} & \cellcolor[HTML]{ECF4FF}46.75 & \cellcolor[HTML]{ECF4FF}77.80 & \multicolumn{1}{c|}{\cellcolor[HTML]{ECF4FF}19.26} & \cellcolor[HTML]{ECF4FF}74.18 & \cellcolor[HTML]{ECF4FF}89.56 & \multicolumn{1}{c|}{\cellcolor[HTML]{ECF4FF}50.78} & \cellcolor[HTML]{ECF4FF}\textbf{72.07} & \cellcolor[HTML]{ECF4FF}\textbf{35.02}  \\ \hline
\end{tabular}
}
\end{center}
\caption{The evaluation results for the proposed unified (U) model with five different compositors on UIGR test split.
For each compositor, the compared model (I) is the combination of two models independently trained on TGR and VCR.
}
\vspace*{-4mm}
\label{tab:main_results}
\end{table}

We compare our multi-task model with previous methods where TGR and VCR are studied independently \footnote{We put implementation details, hyperparameter settings, evaluation protocols, ablation study and qualitative results in Supp. Mat.}. 
The main experiment results are reported in Table \ref{tab:main_results}.
We can draw the following conclusions from the results:
(1) Overall, our proposed multi-task model achieves comparable and even better performance (1.18 mAP increase on average) compared with the combination of two separately trained models.
The best result (the last row) is achieved by our multi-task model with RTIC \cite{shin2021rtic} as the compositor. 
(2) In most cases (4 out of 5), our model achieves significantly better performance than an independently trained model on the VCR task.
It suggests that text is more suitable than one-hot labels as the user feedback for VCR.
With the user feedback in the same modality of TGR, VCR can learn useful information from TGR in our unified model.
(3) Although our model has a slight performance drop on the TGR task, its performance is still competitive against an independently trained model on the TGR subset (\eg, only 0.58 mAP drop for TGR but 2.95 mAP gain for VCR on average).

In summary, the experiment results demonstrate that VCR and TGR can be unified and implemented in a single model through our proposed framework. 
It is more efficient by having one model only and more effective with improved overall performance over the two tasks.

\section{Conclusion}
We have proposed a unified setting for TGR and VCR with a new large-scale benchmark and a baseline multi-task architecture, in which we use text as the unified user feedback format for both TGR and VCR.
We conducted experiments to show that the proposed baseline model has competitive or even better performance than previous methods, and it is also more efficient to use one model instead of two.

\twocolumn[
\begin{@twocolumnfalse}
	\section*{\centering{\vspace{15mm} \\ \Large \textbf{UIGR: Unified Interactive Garment Retrieval \\ \vspace{6mm} -- \\ \vspace{6mm} \large Supplementary Material \\ [40pt]}}}
\end{@twocolumnfalse}
]
\appendix
\section{Additional information on UIGR dataset}
Our dataset is built upon Fashionpedia \cite{jia2020fashionpedia}, which is a large-scale dataset for garment segmentation and fine-grained attribute localization. 
Fashionpedia provides an ontology built by fashion experts containing 27 garment categories and 19 garment parts. 
It provides not only fine-grained attributes but also implicit visual compatibility relationships for all garments in an outfit.
All alternatives \cite{han2017automatic,lin2020fashion} cannot meet all these conditions at the same time.

\vspace*{1mm}
\noindent \textbf{Image pre-processing.}
We want an IGR model to focus on the garment to be refined by the user feedback. 
The background and other garment items in a given image are thus distractions and should be removed.
To this end,  a series of pre-processing steps are introduced:
(1) We use a salient object detection model \cite{qin2019basnet,zhuge2021kaleido} to remove the background, which is an easy task given the typical clean background in fashion catalog images. 
(2) When there are multiple garments with the same category in one image (\eg, shoes and gloves), if they do not overlap, we only keep the one with the largest pixel area; 
(3) We delete the masks of garment parts (\eg, sleeves and pockets) but merge their attributes into the garments they belong to; 
(4) We delete the garments that have low-resolution or extreme aspect ratio;
(5) If there are pixels of other garments in the bounding box, we mask these excess pixels with gray color.
Finally, we cropped each garment with its attributes from the original image to construct a substantial image pool.

\begin{table}[t]
\begin{center}
\resizebox{1\linewidth}{!}{
\begin{tabular}{|c|}
\hline
\textbf{TGR}                                                          \\ \hline
search another item with a similar style                              \\
there are no changes between two images                               \\
change \{$A$\} to \{$V$\}                                                 \\
has \{$V$\} \{$A$\}                                                       \\
is \{$V$\}                                                              \\
change \{$A$\} to \{$V_1$\} and \{$V_2$\}                                 \\
change \{$A_1$\} to \{$V_1$\} and change \{$A_2$\} to \{$V_2$\}           \\
has \{$V_1$\} and \{$V_2$\} \{$A_2$\}                                    \\
has \{$V_1$\} \{$A_1$\} and \{$V_2$\} \{$A_2$\}                           \\
is \{$V_1$\} and with \{$V_2$\} \{$A_2$\}                                \\
is \{$V_1$\} and \{$V_2$\}                                              \\ \hline \hline
\textbf{VCR}                                                          \\ \hline
search a \{$TC$\} that matches this \{$RC$\} best                         \\
retrieve a \{$TC$\} having a similar style with current \{$RC$\}          \\
for this \{$RC$\}, find a visually compatible \{$TC$\}                    \\
replace this \{$RC$\} with a \{$TC$\} that has a consistent style         \\
search a \{$TV$\} \{$TC$\} that matches this \{$RC$\} best                  \\
retrieve a \{$TV$\} \{$TC$\} having a similar style with current \{$RC$\}   \\
for this \{$RC$\}, find a visually compatible \{$TC$\} with \{$TV$\} \{$TA$\} \\
replace this \{$RC$\} with a \{$TC$\} that has \{$TV$\} \{$TA$\}              \\ \hline
\end{tabular}
}
\end{center}
\caption{
All prompts for user feedback generation of UIGR.
\{$V$\} and \{$A$\} hold the blank for one attribute name and its value.
\{$TV$\}, \{$TC$\} and \{$RC$\} stand for the target attribute value, target category and reference category, respectively.
Which kind of TGR prompt to choose depends on how many related attributes (0, 1 or 2) need to be mentioned.
Which kind of VCR prompt to choose depends on whether the target attributes need to be mentioned.
}
\label{tab:prompts}
\end{table}
\vspace*{1mm}
\noindent \textbf{Prompt engineering.}
We list all used prompts for user feedback generation in Table \ref{tab:prompts}.
Our prompts simulate a variety of syntax structures: single phrases, compositional phrases, and propositional phrases.

\begin{figure}[t]
\centering
\includegraphics[width=0.9\linewidth]{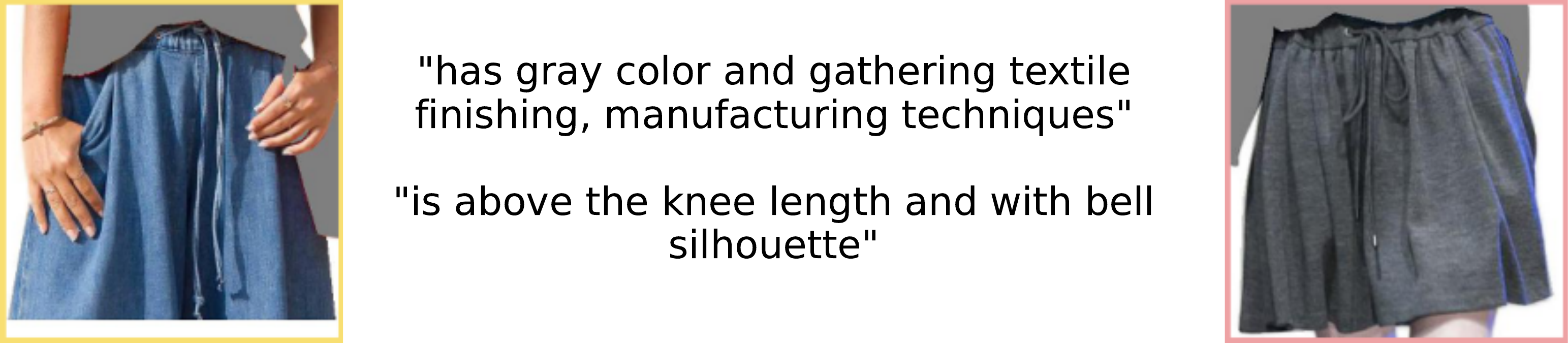}
\includegraphics[width=0.9\linewidth]{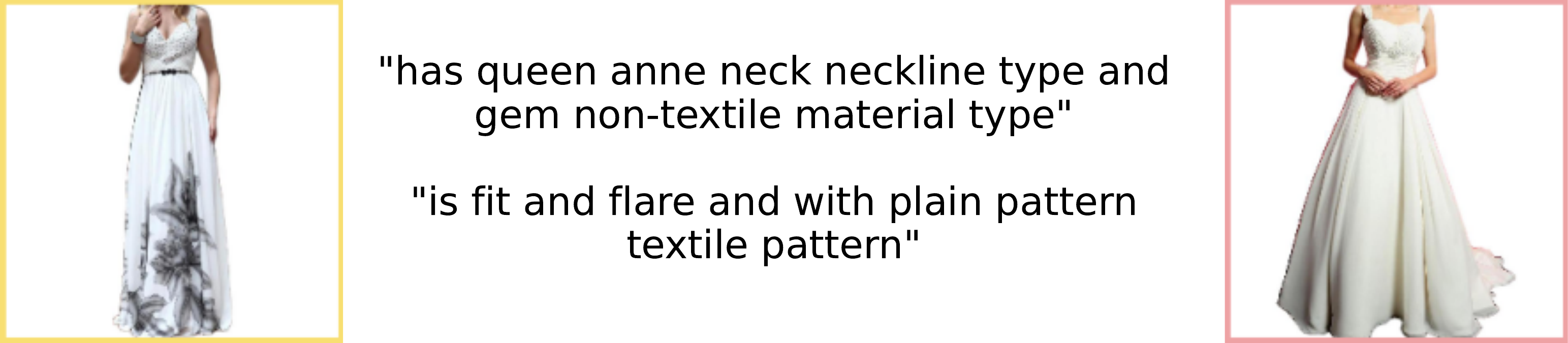}
\includegraphics[width=0.9\linewidth]{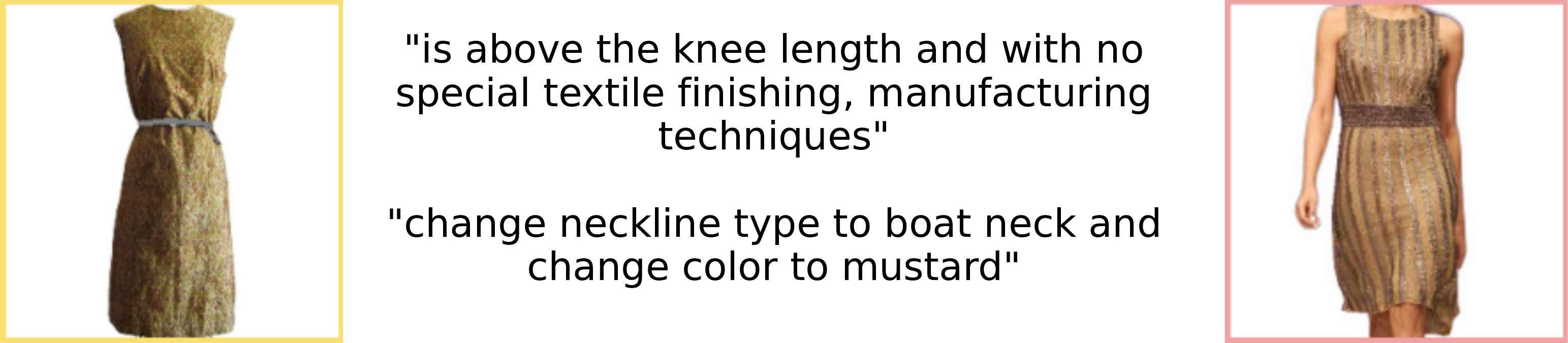}
\includegraphics[width=0.9\linewidth]{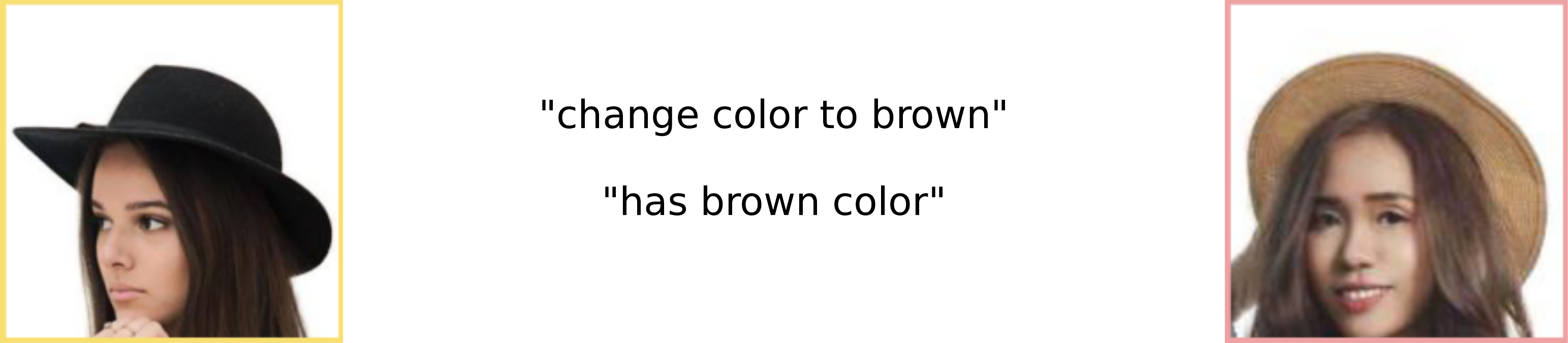}
\includegraphics[width=0.9\linewidth]{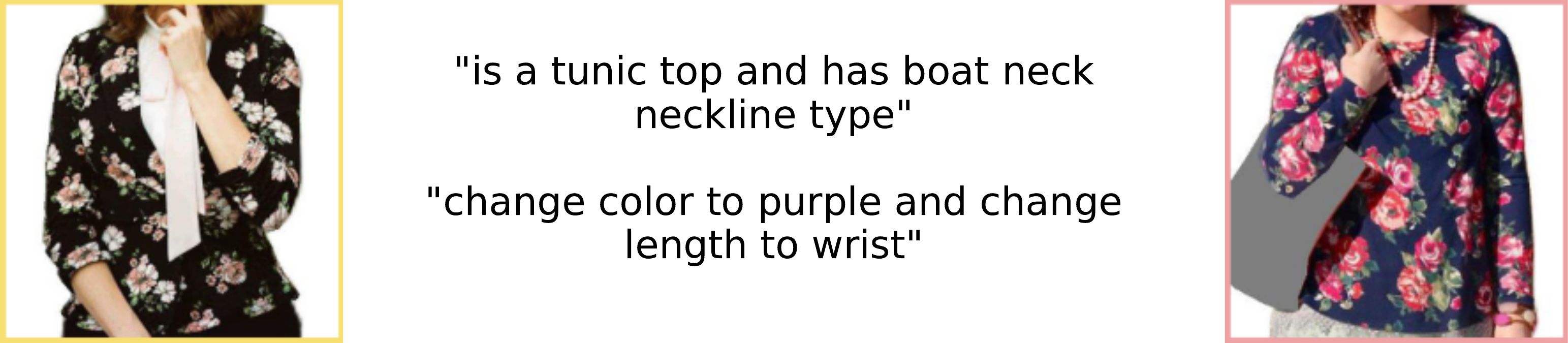}
\includegraphics[width=0.9\linewidth]{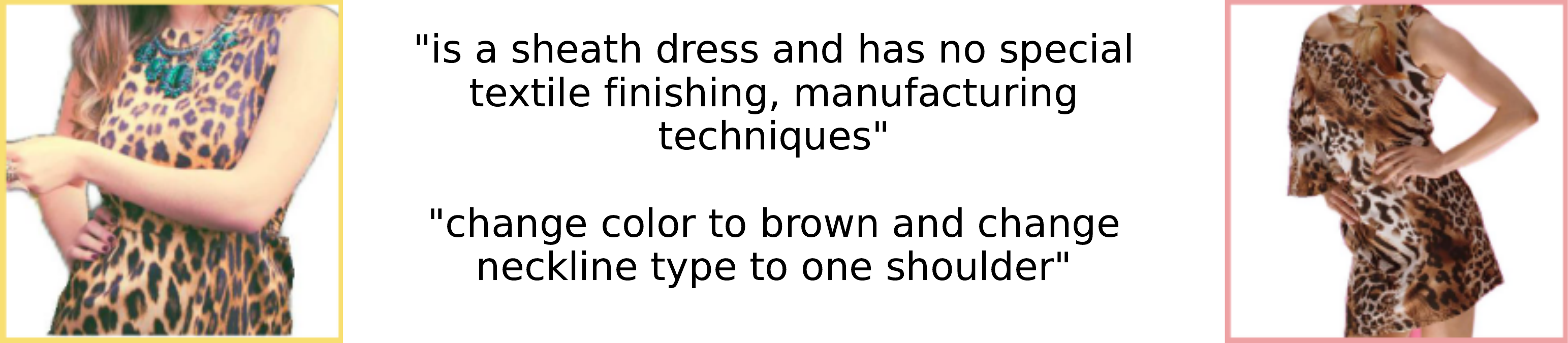}
\includegraphics[width=0.9\linewidth]{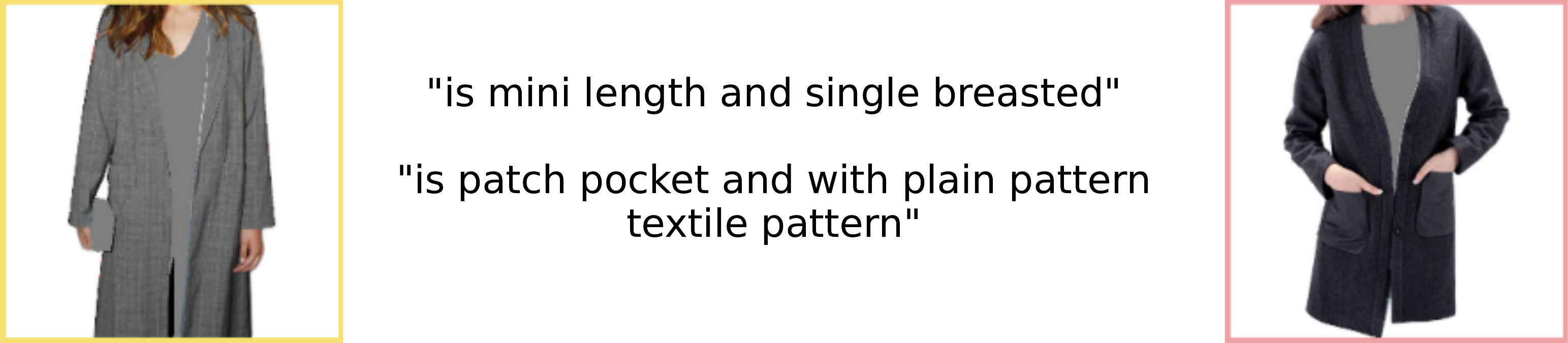}
\includegraphics[width=0.9\linewidth]{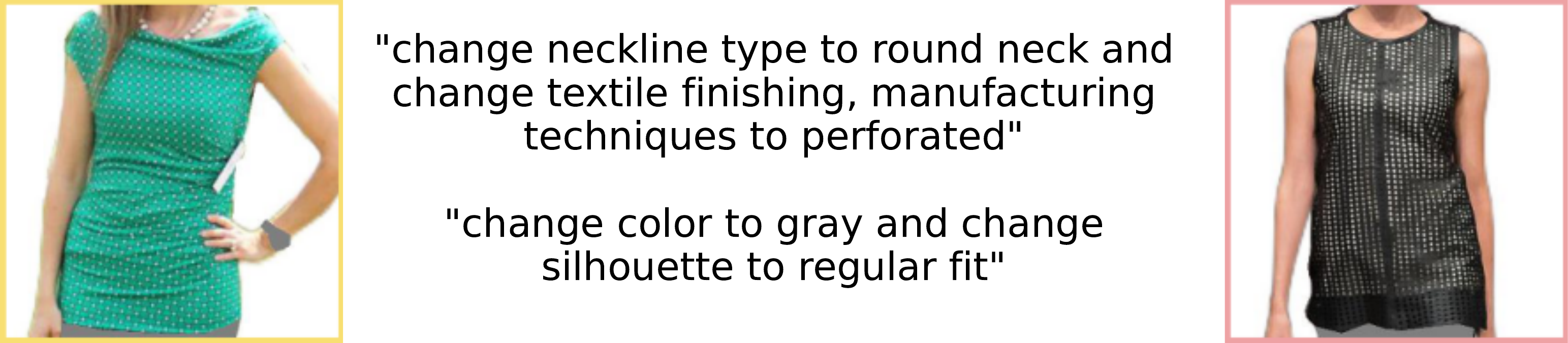}
\includegraphics[width=0.9\linewidth]{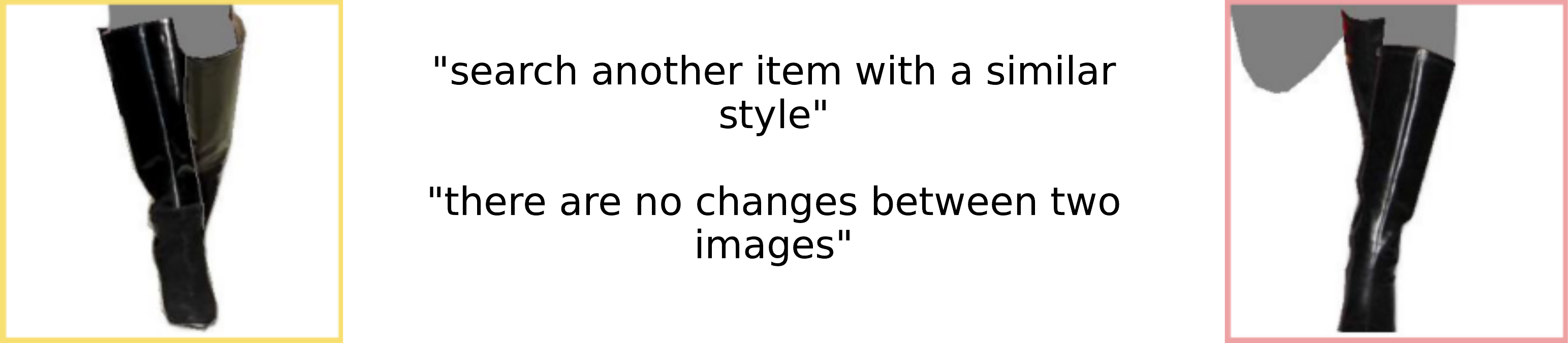}
\caption{More triplet examples in UIGR TGR subset.}
\label{fig:more_tgr_examples}
\end{figure}

\begin{figure}[t]
\centering
\includegraphics[width=0.9\linewidth]{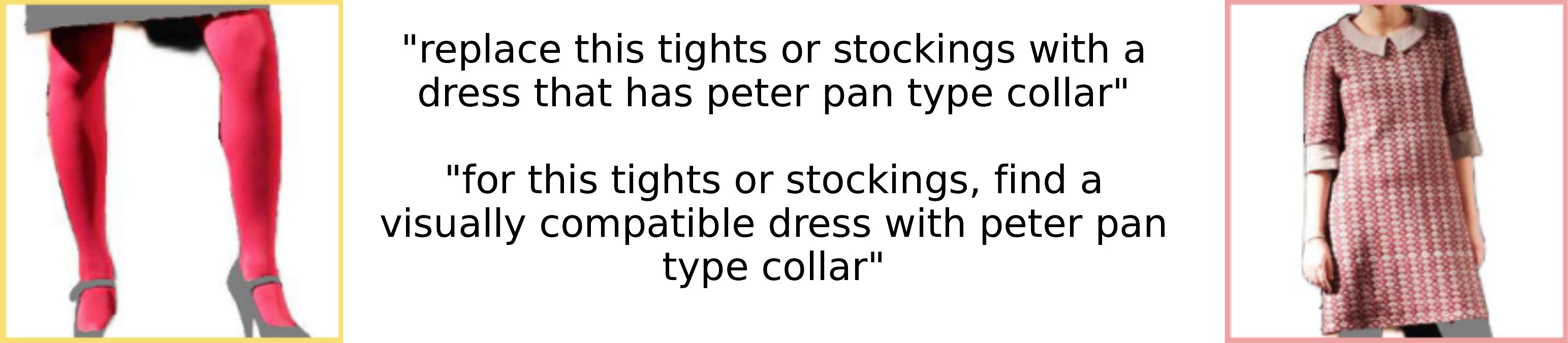}
\includegraphics[width=0.9\linewidth]{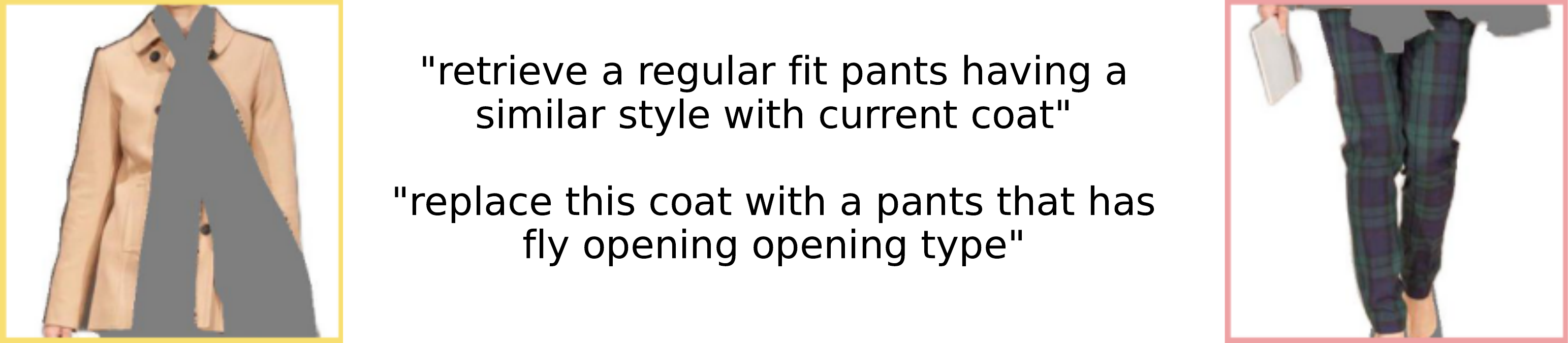}
\includegraphics[width=0.9\linewidth]{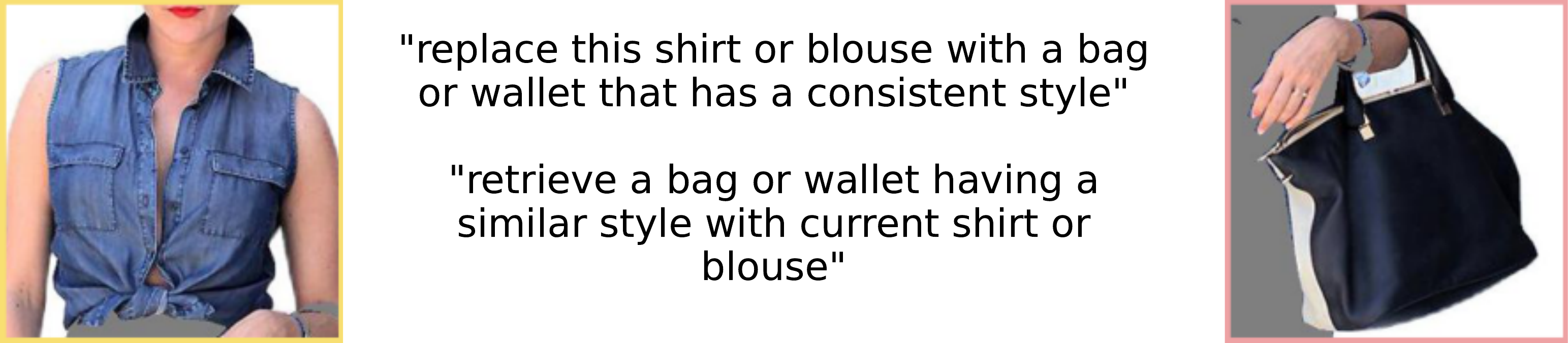}
\includegraphics[width=0.9\linewidth]{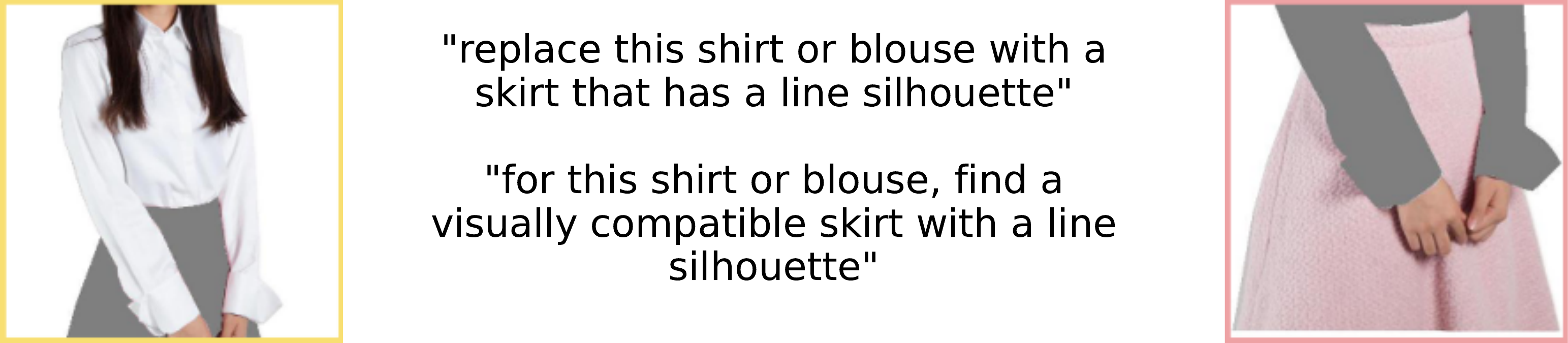}
\includegraphics[width=0.9\linewidth]{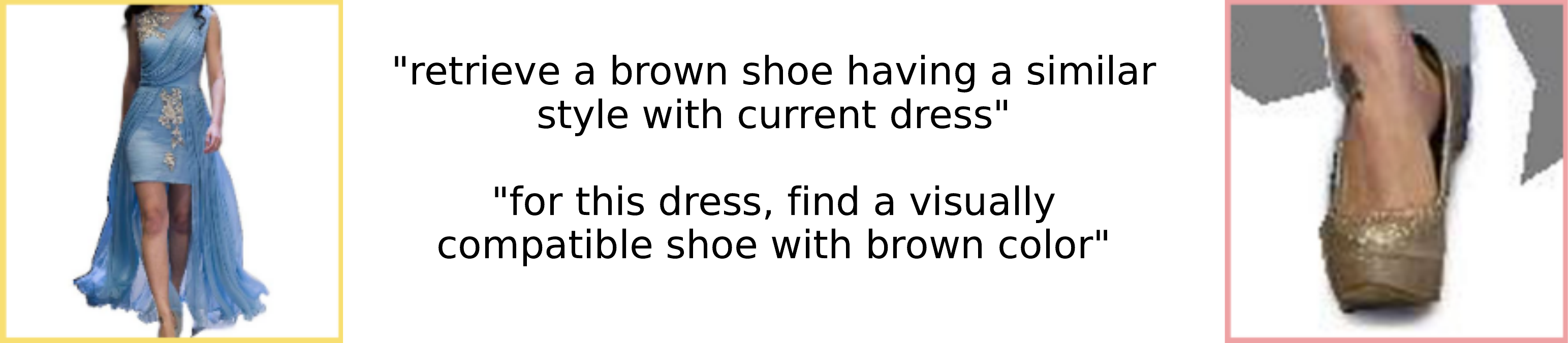}
\includegraphics[width=0.9\linewidth]{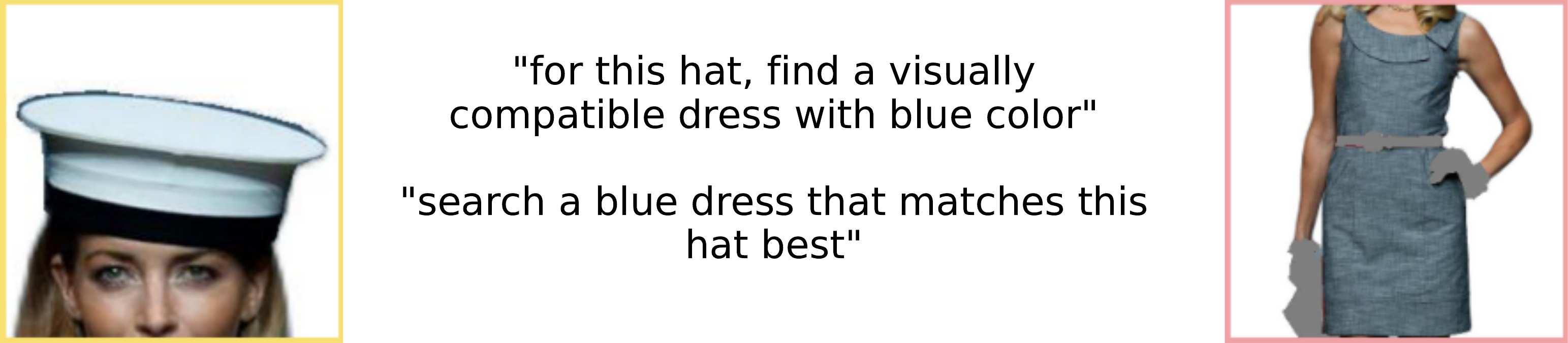}
\includegraphics[width=0.9\linewidth]{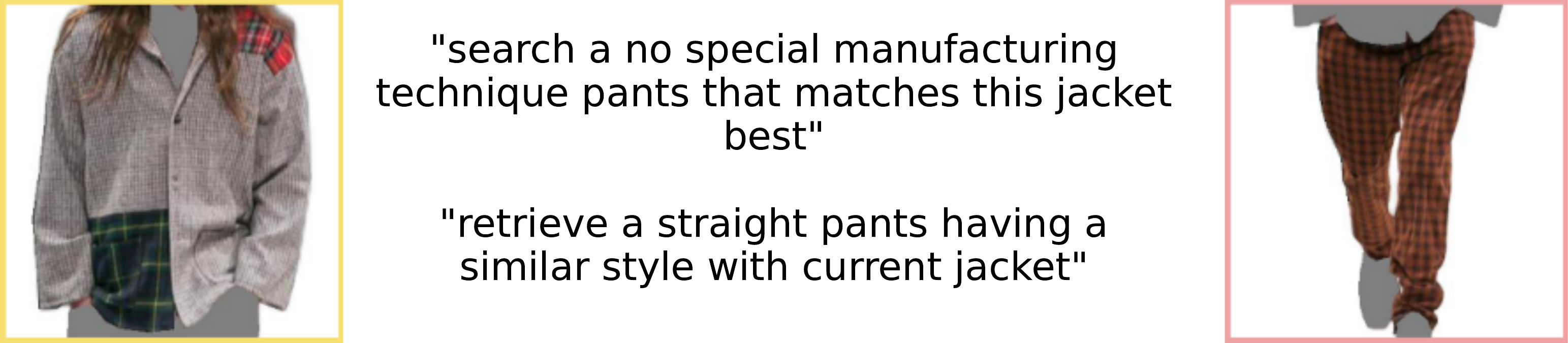}
\includegraphics[width=0.9\linewidth]{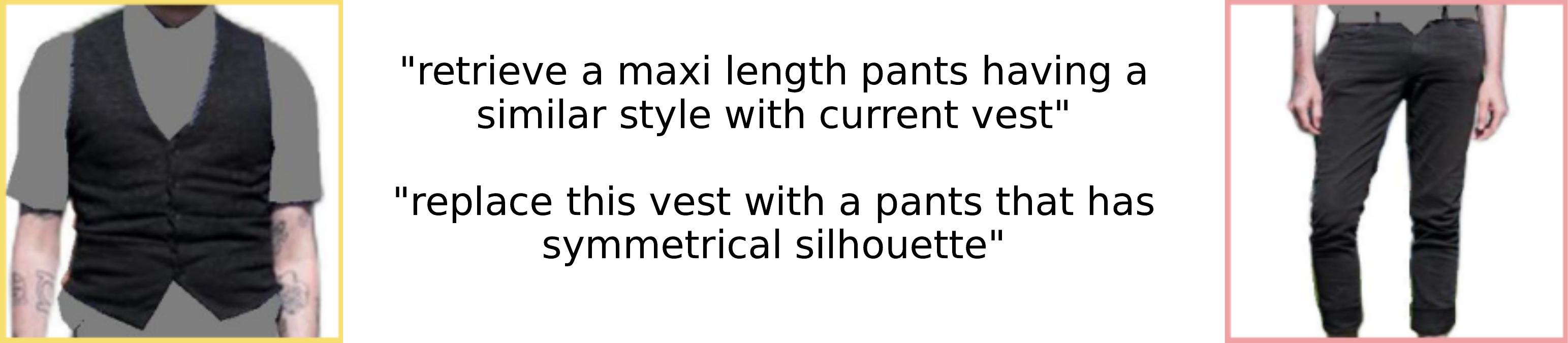}
\includegraphics[width=0.9\linewidth]{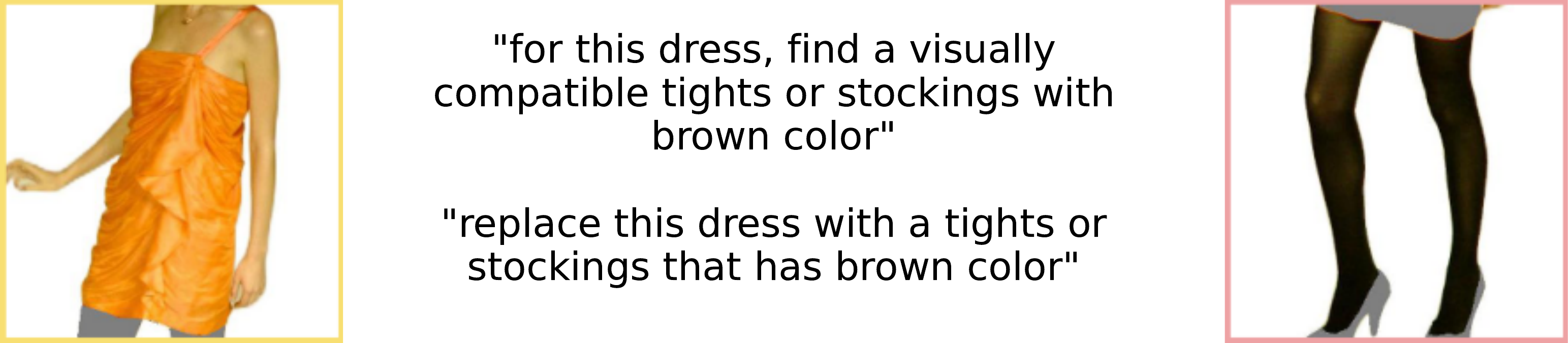}
\caption{More triplet examples in UIGR VCR subset.}
\label{fig:more_vcr_examples}
\end{figure}

\vspace*{1mm}
\noindent \textbf{More triplet examples.}
We present more triplet examples of UIGR in Figure \ref{fig:more_tgr_examples} and Figure \ref{fig:more_vcr_examples}.
As discussed in the main paper, the TGR triplets we collected successfully follow the assumption that there could not be too many visual changes between the reference garment and the target garment.
Our TGR triplets thus are much higher quality than those in FashionIQ \cite{wu2021fashioniq} with less ambiguity.
Besides, our TGR subset contains 27 different garments, far more than FashionIQ, which only has three categories (top tee, shirt and dress).

Thanks to the flexibility of text, our VCR subset includes more
meaningful information compared to concatenated one-hot labels \cite{lin2020fashion}.
Now each user feedback sentence states category changes and intended attributes based on the statistics of attribute co-occurrence between compatible garment items, which is more in line with reality.
With the help of such kind of VCR triplets, the potential user can specific the search direction through mentioning some specific target attributes.
Most importantly, now the VCR has the unified setting with TGR.

\section{Additional information on the multi-task baseline model}
Given a reference garment image $g^r$ and an interaction signal (user feedback) $s$, the ultimate goal of interactive retrieval is to search the gallery for another garment image $g^t$ that best matches the modification mentioned in $s$.
Regardless of whether the user wants to modify the attributes or the category of the reference garment, the interaction signal is in the same textual format.
TGR and VCR can thus be modeled in the same framework.

We will first briefly introduce how previous works study these two tasks separately and then describe our unified solution based on multi-task learning.

\subsection{Preliminary method}
\begin{figure}[t]
\begin{center}
\includegraphics[width=0.85\linewidth]{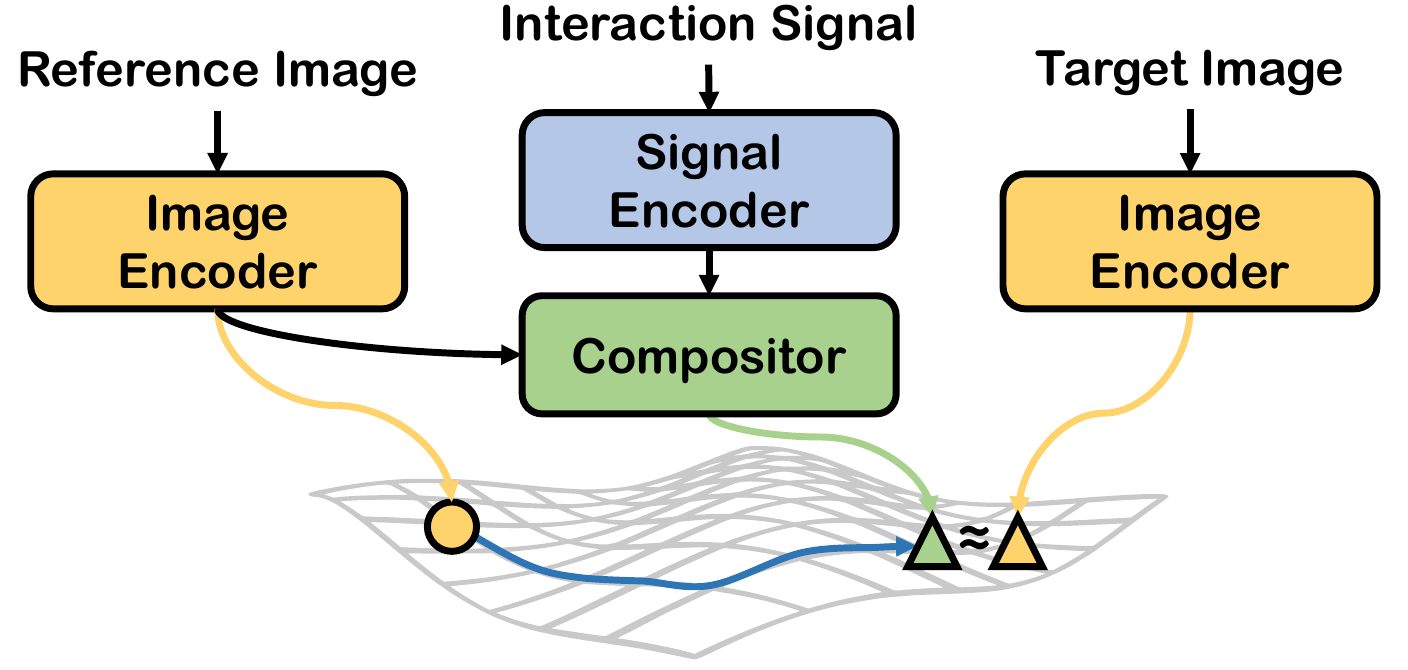}
\end{center}
\vspace*{-3mm}
\caption{Previous architecture for TGR/VCR.}
\label{fig:previous_pipeline}
\vspace*{-2mm}
\end{figure}

In Figure \ref{fig:previous_pipeline}, an existing pipeline for interactive retrieval typically consists of three components: image encoder $\mathcal{E}^I$, interaction signal encoder $\mathcal{E}^S$ and compositor $\mathcal{C}$.

Firstly, both reference image and target image are fed into the image encoder to obtain representations in the feature space:
$
\noindent \textbf{g}^r = \mathcal{E}^I\left(g^r\right),
\noindent \textbf{g}^t = \mathcal{E}^I\left(g^t\right),
$
where $\mathcal{E}^I$ is usually instantiated by a CNN pre-trained on ImageNet \cite{deng2009imagenet} and a linear projection layer \cite{vo2019tirg,shin2021rtic}.

In the meantime, the interaction signal is processed by the signal encoder to get the signal representation:
$
\noindent \textbf{s} = \mathcal{E}^S\left(s\right),
$
where the interaction signal is represented by the concatenation of reference category $c^r$ and target category $c^t$ for VCR \cite{lin2020fashion} or by user feedback $t$ for TGR \cite{vo2019tirg}.

Finally, the most important step is to incorporate the interaction signal's feature into reference image's feature via a compositor:
$
\noindent \textbf{x} = \mathcal{C}\left(\noindent \textbf{g}^r, \noindent \textbf{s}\right).
$
For VCR, this compositor is always instantiated by a conditional similarity module \cite{lin2020fashion,hou2021disentangled} to learn different sub-spaces with different notions.
For TGR, this compositor works globally \cite{vo2019tirg,shin2021rtic} or locally \cite{chen2020val,lee2021cosmo} to modify the feature map of reference image.

The goal of this pipeline is to make the composed query $\noindent \textbf{x}$ as close as possible to the target $\noindent \textbf{g}^t$ in a shared feature space.
A widely used objective function is the batch-based classification loss (BBC) \cite{vo2019tirg}, which assumes the same form as the InfoNCE loss \cite{oord2018infonce}:
\begin{equation}
\mathcal{L}_{bbc}=\frac{1}{B} \sum_{i=1}^{B}-\log \frac{\exp \left[\kappa\left(\noindent \textbf{x}_{i}, \noindent \textbf{g}^t_i\right) / \tau \right]}{\sum_{j=1}^{B} \exp \left[\kappa\left(\noindent \textbf{x}_{i}, \noindent \textbf{g}^t_j\right) / \tau \right]},
\label{eqa:bbc_loss}
\end{equation}
where $\kappa(\cdot,\cdot)$ and $\tau$ are cosine distance metric and tuneable temperature, respectively.
In this loss, each example is contrasted with a set of other negatives.
It thus achieves better discriminative learning and faster convergence.

During inference, the features of all gallery images will be calculated in advance by image encoder.
For each composed query, its cosine similarity with all gallery features will be obtained.
Finally, an identity list is sorted according to the cosine similarity as the retrieval result sequence.

\begin{figure}[t]
\begin{center}
\includegraphics[width=\linewidth]{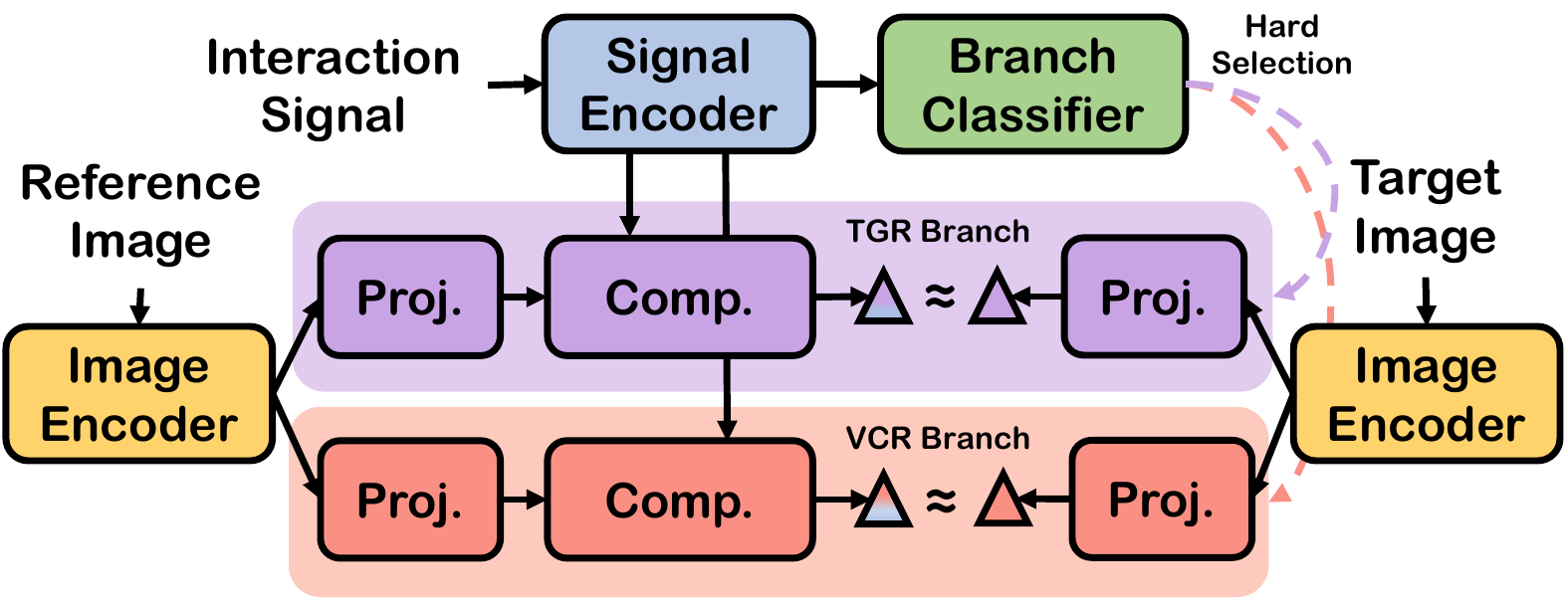}
\end{center}
\vspace*{-3mm}
\caption{Proposed multi-task architecture for UIGR.}
\label{fig:proposed_pipeline}
\vspace*{-4mm}
\end{figure}

\subsection{Proposed multi-task framework}
Although there are different implementations for the compositors of VCR and TGR, they share the same goal: preserving unmentioned visual appearance aspects of the reference and changing only those mentioned in the interaction signal/feedback.
Our multi-task model unifies the two tasks based on the same goal. 
However, to accommodate the major difference in the change directions of the two tasks, namely whether the category is preserved or changed, we use different compositors. 
As shown in Figure \ref{fig:proposed_pipeline}, two branches are used for separately learning two composition processes with shared image and signal encoders.

More specifically,  we use a quintuplet $\left\{g^r, s_v, s_t, g^t_v, g^t_t \right\}$ containing reference garment, VCR signal, TGR signal, VCR target garment, and TGR target garment as the input for training.
These three garment images will be fed into a shared image encoder $\mathcal{E}^I$ to get respective features $\noindent \textbf{g}^r$, $\noindent \textbf{g}^t_v$ and $\noindent \textbf{g}^t_t$.
Similarly, two signals will get their features $\noindent \textbf{s}_v$ and $\noindent \textbf{s}_t$ via a shared signal encoder $\mathcal{E}^S$.

Considering that the features needed to be modified for the two branches are not the same, we use two projection modules $\mathcal{P}_t$  and $\mathcal{P}_v$ to project image features to two latent spaces ahead of the composition process. Exactly how the projection module is realized depends on what compositor is employed here. 
Specifically, for the compositor who directly modifies the feature map \cite{chen2020val,lee2021cosmo}, we implement the projection module with a lightweight CNN; for the compositor working globally \cite{vo2019tirg,shin2021rtic}, we use a linear projection layer following the global average pooling instead.

After choosing a compositor architecture from a existing method (\eg, \cite{vo2019tirg,shin2021rtic, chen2020val,lee2021cosmo}), we need two compositors $\mathcal{C}_t$ and $\mathcal{C}_v$ of the same architecture but without shared weights, to separately learn two composition processes for the two tasks.
For each branch, the compositor serves for incorporating signal feature into the projected image feature of reference garment: 
\begin{equation}
\noindent \textbf{x}_v = \mathcal{C}_v\left(\mathcal{P}_v\left(\noindent \textbf{g}^r\right), \noindent \textbf{s}_v\right),
\quad
\noindent \textbf{x}_t = \mathcal{C}_t\left(\mathcal{P}_t\left(\noindent \textbf{g}^r\right),\noindent \textbf{s}_t\right).
\end{equation}
For both branches, two BBC losses $\mathcal{L}_{bbc}^v$ and $\mathcal{L}_{bbc}^t$ will be calculated independently according to Equation \ref{eqa:bbc_loss}.

We also jointly learn a classifier to distinguish different user feedback.
Specifically, we simply choose the branch with a higher score predicted by the classifier, \ie, hard selection, which is empirically found to be the most effective design.
We instantiate this branch classifier with an MLP $\mathcal{M}$ and optimize it via cross-entropy loss (CE):
\begin{equation}
\small
\begin{aligned}
\mathcal{L}_{ce}=\frac{1}{B} \sum_{i=1}^{B}-\log \frac{\exp \left[\mathcal{M}_0 \left(\noindent \textbf{s}_{vi}\right) \right]}{\exp\left[\mathcal{M}_0 \left(\noindent \textbf{s}_{vi}\right) \right] + \exp\left[\mathcal{M}_1 \left(\noindent \textbf{s}_{vi}\right) \right]}
\\
+\frac{1}{B} \sum_{i=1}^{B}-\log \frac{\exp \left[\mathcal{M}_1 \left(\noindent \textbf{s}_{ti}\right) \right]}{\exp\left[\mathcal{M}_0 \left(\noindent \textbf{s}_{ti}\right) \right] + \exp\left[\mathcal{M}_1 \left(\noindent \textbf{s}_{ti}\right) \right]}.
\end{aligned}
\end{equation}

Our model is end-to-end optimized by the overall objective function, which is the direct summation of two BBC losses and one CE loss:
\begin{equation}
    \mathcal{L} = \mathcal{L}_{bbc}^v + \mathcal{L}_{bbc}^t + \mathcal{L}_{ce}.
\end{equation}

\begin{table}[t]
\begin{center}
\resizebox{\linewidth}{!}{
\begin{tabular}{|c|ccc|ccc|cc|}
\hline
\multirow{2}{*}{\textbf{Arch.}} & \multicolumn{3}{c|}{\textbf{TGR Results}}    & \multicolumn{3}{c|}{\textbf{VCR Results}}    & \multicolumn{2}{c|}{\textbf{Mean}}              \\ \cline{2-9} 
                                & \textbf{R@10} & \textbf{R@50} & \textbf{mAP} & \textbf{R@10} & \textbf{R@50} & \textbf{mAP} & \textbf{R@K}           & \textbf{mAP}           \\ \hline
\textbf{I}                      & 46.27         & 77.57         & 19.78        & 69.30         & 85.88         & 46.15        & 69.76                  & 32.97                  \\ \hline
\textbf{U+SC}                   & 43.97         & 76.22         & 17.67        & 71.18         & 87.89         & 46.89        & 69.82                  & 32.28                  \\
\textbf{U+SP}                   & 42.74         & 75.04         & 17.84        & 69.67         & 87.51         & 45.60        & 68.74                  & 31.72                  \\
\textbf{U+SC+SP}                & 43.94         & 75.76         & 18.30        & 68.83         & 87.26         & 44.42        & 68.95                  & 31.36                  \\ \hline
\rowcolor[HTML]{ECF4FF} 
\textbf{U}                      & 45.10         & 76.84         & 18.94        & 72.15         & 88.61         & 48.49        & \textbf{70.68}         & \textbf{33.72}         \\\hline
\end{tabular}
}
\end{center}
\caption{Ablation study on the proposed multi-task model. SC: sharing compositor across two branches; SP: sharing projection module across two branches.}
\label{tab:ablation_study}
\end{table}
\begin{table}[t]
\begin{center}
\resizebox{\linewidth}{!}{
\begin{tabular}{|c|c|c|c|c|}
\hline
\textbf{User Feedback}    & \textbf{Attribute Augmented} & \textbf{R@10} & \textbf{R@50} & \textbf{mAP} \\ \hline
\textbf{One-hot} &               & 69.30         & 85.88         & 46.15        \\ \hline
\textbf{One-hot} & \cmark           & 70.98         & 87.16         & 47.80        \\ \hline
\textbf{Text}    &               & 70.77         & 86.88         & 47.51        \\ \hline
\textbf{Text}    & \cmark           & \textbf{72.65}         & \textbf{88.64}         & \textbf{49.06}        \\ \hline
\end{tabular}
}
\end{center}
\caption{Experiment results of attribute argumented models (with one-hot labels or text as the user feedback) on VCR subset.}
\label{tab:vcr_more}
\end{table}

\section{Additional information on experiments}
\noindent \textbf{Implementation details.}
We realize the image encoder and signal encoder by utilizing ResNet50 \cite{he2016resnet} and Bi-GRU \cite{cho2014gru}.
The ResNet50 is pre-trained on ImageNet \cite{deng2009imagenet} and the word embeddings of Bi-GRU are initialized by CLIP text encoder \cite{radford2021clip, han2021tbps}.
To demonstrate the universality of our multi-task architecture, we instantiate the compositor with recent representative methods \cite{lin2020fashion,vo2019tirg,chen2020val,lee2021cosmo,shin2021rtic}.
For the projection module, we adopt two different architectures (convolution layer with 512 output channels or linear layer with 512 output dimensions) according to whether the compositor is used to modify the feature map or the global feature.

\vspace{1mm}
\noindent \textbf{Hyper-parameters setting.}
We use random horizontally flip and random crop as image data augmentation methods.
All images are resized to $224 \times 224$. 
The batch size and temperature in the $\mathcal{L}_{bbc}$ are 64 and 0.0625, respectively.
Our model is trained with Adam optimizer \cite{kingma2014adam} for 40 epochs with an initial learning rate $2 \times 10^{-4}$, which is decayed by a factor 0.1 at the $15^{th}$ and $25^{th}$ epoch, respectively. 
We also linearly increase the learning rate from $2 \times 10^{-5}$ to $2 \times 10^{-4}$ at the first 5 epochs.
All experiments are conducted on one Tesla V100 GPU (32GB memory) with Pytorch \cite{paszke2019pytorch}.

\vspace{1mm}
\noindent \textbf{Evaluation metrics.}
We adopt the standard evaluation metric for retrieval, \ie, Recall@K, denoted as R@K for short.
To circumvent the problem of false negatives \cite{liu2021cirr}, we follow FashionIQ \cite{wu2021fashioniq} to set K as larger values (10 and 50).
In addition, we also report the mean Average Precision (mAP) \footnote{For each query, mAP is calculated with top 50 results.} for a comprehensive evaluation. 

\vspace{1mm}
\noindent \textbf{Evaluation protocols.}
Since we are integrating VCR into TGR, we want the model has the ability to distinguish different categories.
Consequently, we lead a more difficult evaluation protocol than FashionIQ.
Unlike FashionIQ, which evaluates three categories separately, category labels are not available for our evaluation protocol.
That is, all images in the gallery will calculate a similarity with the composed query.

\section{More quantitative results}
\subsection{Ablation study}
We examine the design of each component in our proposed model. 
The critical problem we are going to explore is whether compositor and projection module can be shared between TGR and VCR.
In all experiments, we remove the branch classifier and use TIRG \cite{vo2019tirg} as the compositor.

As shown in Table \ref{tab:ablation_study}, sharing both projection module and compositor leads to a performance drop. 
In addition, a shared projection module alone leads to a more considerable performance drop than a shared compositor.
This result demonstrates that projecting the features of reference garments into different latent spaces is vital for this multi-task framework. To unify VCR and TGR in a single model, the projection module and compositor thus cannot be shared because different tasks need different embedding features.

\subsection{Attribute augmented VCR model}
In addition to helping to unify VCR and TGR, we believe that mentioning target attributes is a more general way for VCR, even for models that use one-hot labels as user feedback.
To demonstrate that, we conduct a small experiment by concatenating the one-hot label of the target attribute behind that of the reference category and target category.

\begin{table*}[ht]
\begin{center}
\resizebox{\linewidth}{!}{
\begin{tabular}{|c|c|ccc|ccc|ccc|cc|}
\hline
\multirow{2}{*}{\textbf{Comp.}} & \multirow{2}{*}{\textbf{Training Dataset}} & \multicolumn{3}{c|}{\textbf{Dress}}              & \multicolumn{3}{c|}{\textbf{Shirt}}              & \multicolumn{3}{c|}{\textbf{Top Tee}}            & \multicolumn{2}{c|}{\textbf{Mean}} \\ \cline{3-13} 
                                &                                            & \textbf{R@10}  & \textbf{R@50}  & \textbf{mAP}   & \textbf{R@10}  & \textbf{R@50}  & \textbf{mAP}   & \textbf{R@10}  & \textbf{R@50}  & \textbf{mAP}   & \textbf{R@K}     & \textbf{mAP}    \\ \hline
\multirow{3}{*}{\textbf{TIRG}\cite{vo2019tirg}}
& UIGR                                       & 7.59           & 19.98          & 3.25           & 7.90           & 18.99          & 3.25           & 8.77           & 23.56          & 3.91           & 14.47            & 3.47            \\
                                & FashionIQ                                  & 23.65          & 49.93          & 11.89          & 21.98          & \textbf{46.61} & 9.31           & 27.84          & 55.07          & 12.53          & 37.51            & 11.24           \\
                                & UIGR + FashionIQ                           & \textbf{26.97} & \textbf{53.64} & \textbf{12.65} & \textbf{22.87} & 46.07          & \textbf{10.29} & \textbf{29.58} & \textbf{57.73} & \textbf{13.80} & \textbf{39.48}   & \textbf{12.25}  \\ \hline
\multirow{3}{*}{\textbf{VAL}\cite{chen2020val}}   & UIGR                                       & 6.05           & 18.20          & 2.78           & 7.31           & 17.76          & 2.85           & 7.50           & 20.04          & 3.05           & 12.81            & 2.89            \\
                                & FashionIQ                                  & 19.09          & 44.57          & 9.02           & 16.68          & 37.93          & 7.21           & 20.45          & 46.76          & 8.88           & 30.91            & 8.37            \\
                                & UIGR + FashionIQ                           & \textbf{26.43} & \textbf{52.66} & \textbf{13.02} & \textbf{20.36} & \textbf{43.52} & \textbf{9.54}  & \textbf{25.85} & \textbf{53.14} & \textbf{12.21} & \textbf{36.99}   & \textbf{11.59}  \\ \hline
\multirow{3}{*}{\textbf{CoSMo}\cite{lee2021cosmo}} & UIGR                                       & 7.14           & 18.80          & 3.23           & 6.04           & 17.52          & 2.73           & 7.45           & 20.96          & 3.22           & 12.99            & 3.06            \\
                                & FashionIQ                                  & 20.87          & 46.80          & 9.35           & \textbf{18.30} & 40.92          & 8.00           & 22.95          & 50.33          & 10.36          & 33.36            & 9.24            \\
                                & UIGR + FashionIQ                           & \textbf{23.50} & \textbf{49.48} & \textbf{10.42} & 17.96          & \textbf{41.76} & \textbf{8.22}  & \textbf{25.14} & \textbf{52.58} & \textbf{11.68} & \textbf{35.07}   & \textbf{10.11}  \\ \hline
\multirow{3}{*}{\textbf{RTIC}\cite{shin2021rtic}}  & UIGR                                       & 8.13           & 21.32          & 3.53           & 7.85           & 20.31          & 3.32           & 9.43           & 23.56          & 4.12           & 15.10            & 3.66            \\
                                & FashionIQ                                  & 25.93          & 51.76          & 12.00          & 22.37          & 46.57          & 9.91           & 27.84          & 56.65          & 13.10          & 38.52            & 11.00           \\
                                & UIGR + FashionIQ                           & \textbf{28.01} & \textbf{53.74} & \textbf{13.58} & \textbf{24.04} & \textbf{47.64} & \textbf{11.36} & \textbf{31.67} & \textbf{57.78} & \textbf{14.92} & \textbf{40.48}   & \textbf{13.29}  \\ \hline
\end{tabular}
}
\end{center}
\caption{The cross-domain (UIGR-TGR $\rightarrow$ FashionIQ \cite{wu2021fashioniq}) evaluation results.
All results are reported on the three subsets of FashionIQ.}
\label{tab:cross_domain}
\end{table*}
As shown in Table \ref{tab:vcr_more}, we can conclude that one-hot labels also benefit from mentioning target attributes, but text modality can integrate this kind of attribute information into user feedback better.

\subsection{Cross-domain evaluation}
To demonstrate the universality of our generated user feedback, we conduct cross-domain evaluation.
Precisely, we compare the results of the same model with 3 different strategies: (1) trained on UIGR-TGR, tested on FashionIQ (zero-shot); trained and tested on FahsionIQ (fully supervised); (3) trained on UIGR-TGR and FashionIQ, and then tested on FashionIQ (transfer learning).
As shown in Table \ref{tab:cross_domain}, we can draw several conclusions: (1) Even under the zero-shot setting, every method achieves reasonable performance; (2) With the transferred knowledge from UIGR-TGR, every model has a substantial performance gain (2.93 R@K and 1.85 mAP increase on average).
In general, although our user feedback is generated, its generalization ability is sufficient to help the model achieve good performance on the manually annotated dataset.

\section{More qualitative results}
\subsection{Visualizations of retrieval results}
\begin{figure}[t]
\centering
\subfigure[is \underline{scoop neck} and change \underline{color} to \underline{mustard}]{
\includegraphics[width=\linewidth]{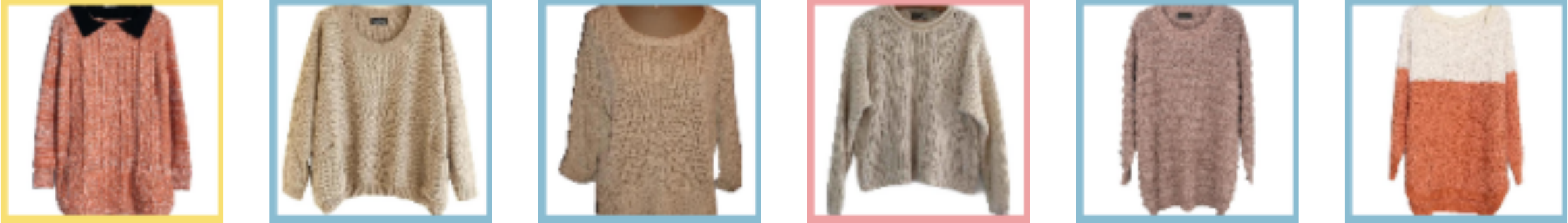}
}
\subfigure[
is a \underline{crop top} and has \underline{round neck neckline type}]{
\includegraphics[width=\linewidth]{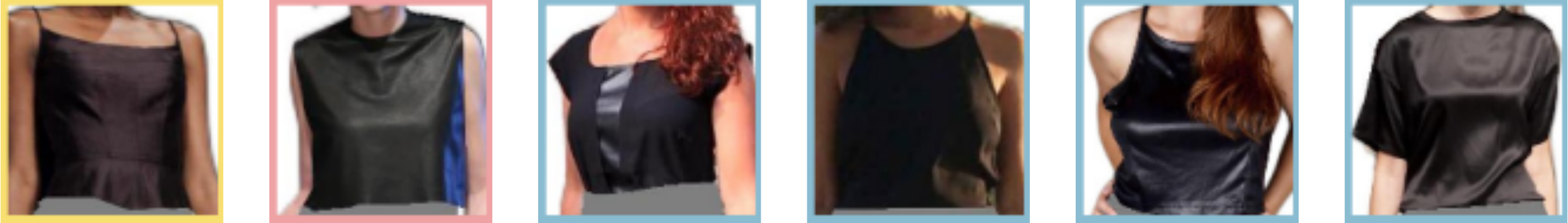}
}
\subfigure[is \underline{plunging neckline} and with \underline{zip up opening type}]{
\includegraphics[width=\linewidth]{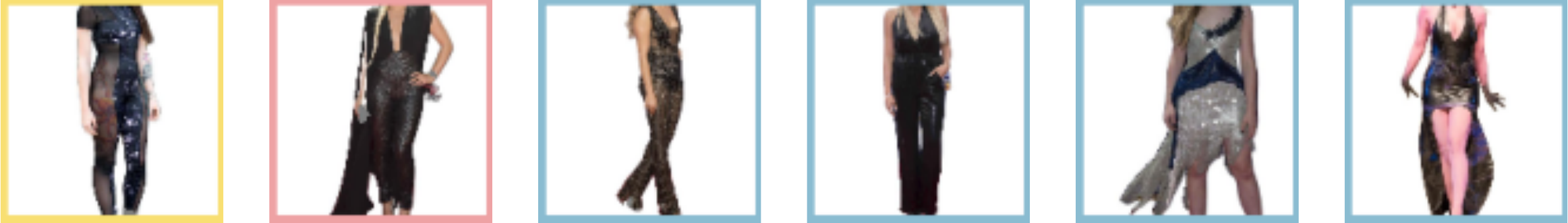}
}
\subfigure[has \underline{flap type pocket} and change \underline{color} to \underline{maroon}]{
\includegraphics[width=\linewidth]{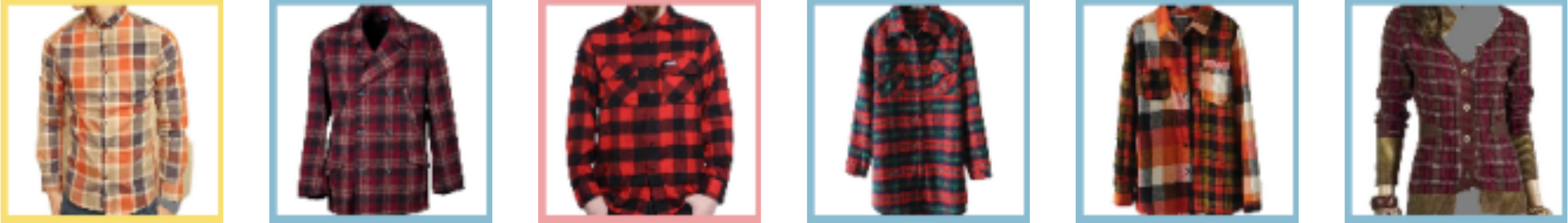}
}
\subfigure[is \underline{applique} and with \underline{loose fit silhouette}]{
\includegraphics[width=\linewidth]{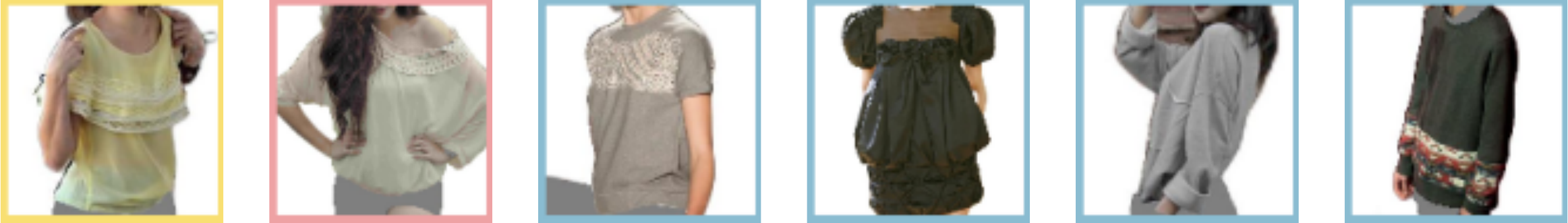}
}
\caption{Retrieval results of our multi-task model on TGR subset.\textcolor{Dandelion}{Yellow}: reference garment; \textcolor{OrangeRed}{Red}: target garment (ground truth); \textcolor{Cerulean}{Blue}: other retrieved garments.}
\label{fig:more_tgr_results}
\end{figure}

\begin{figure}[t]
\centering
\subfigure[search a \underline{brown shoe} that matches this \underline{jacket} best]{
\includegraphics[width=\linewidth]{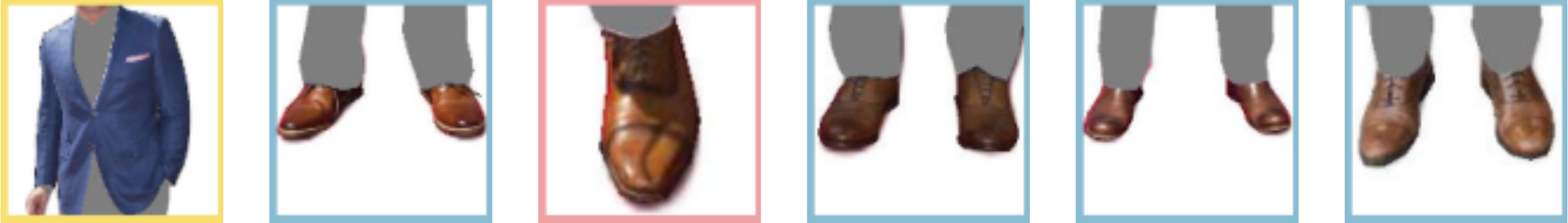}
}
\subfigure[retrieve a \underline{hat} having a similar style with current \underline{skirt}]{
\includegraphics[width=\linewidth]{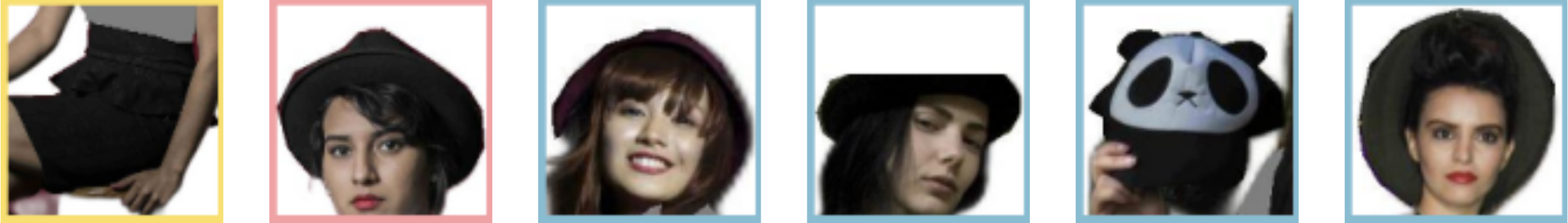}
}
\subfigure[retrieve a \underline{yellow shoe} having a similar style with \underline{current skirt}]{
\includegraphics[width=\linewidth]{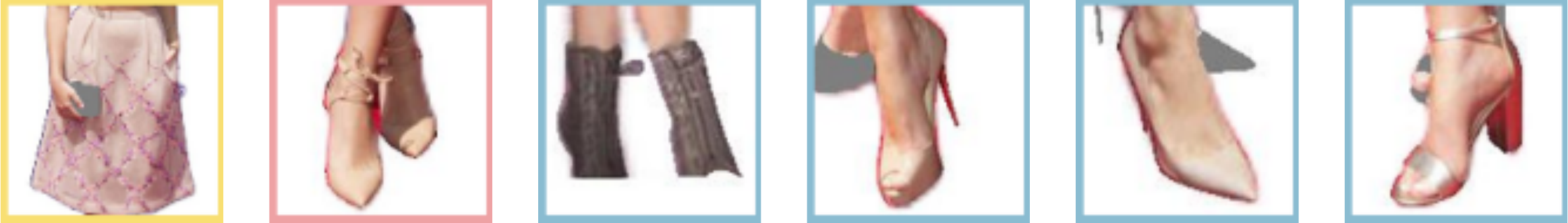}
}
\subfigure[search an \underline{above the hip length jacket} that matches this \underline{shoe} best]{
\includegraphics[width=\linewidth]{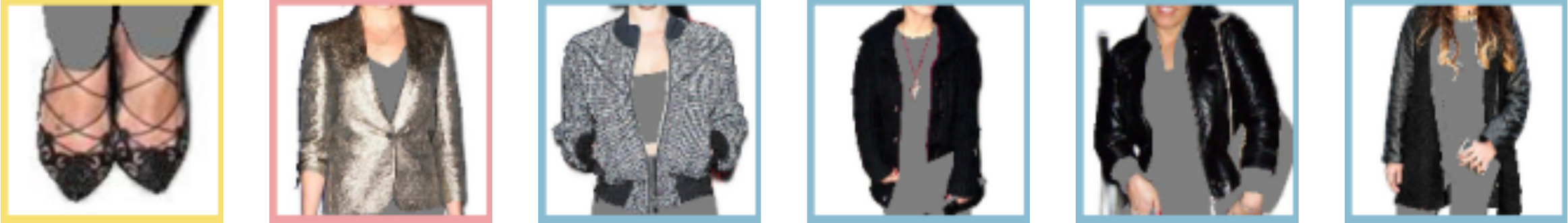}
}
\subfigure[retrieve a \underline{zip up skirt} having a similar style with current \underline{top}]{
\includegraphics[width=\linewidth]{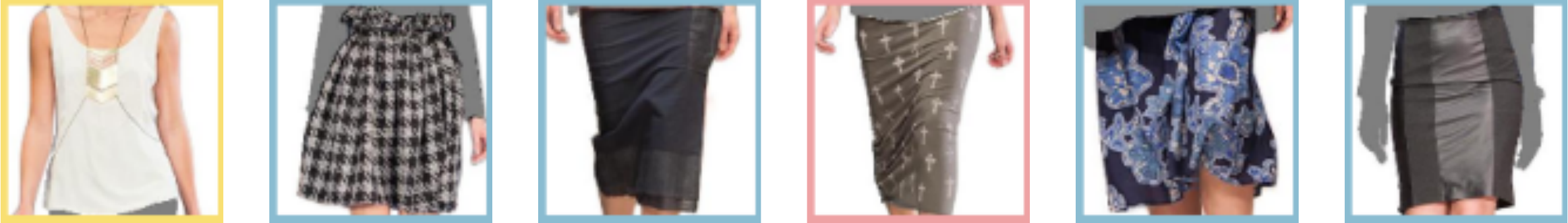}
}
\caption{Retrieval results of our model on VCR subset.}
\label{fig:more_vcr_results}
\end{figure}

\begin{figure}[t]
\centering
\subfigure[is a \underline{sheath dress} and has \underline{printed textile techniques}]{
\includegraphics[width=\linewidth]{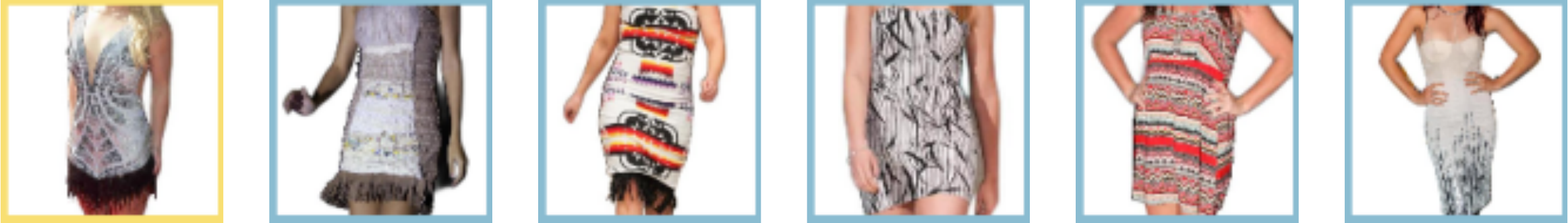}
}
\subfigure[is \underline{napoleon lapel} and \underline{no special manufacturing technique}]{
\includegraphics[width=\linewidth]{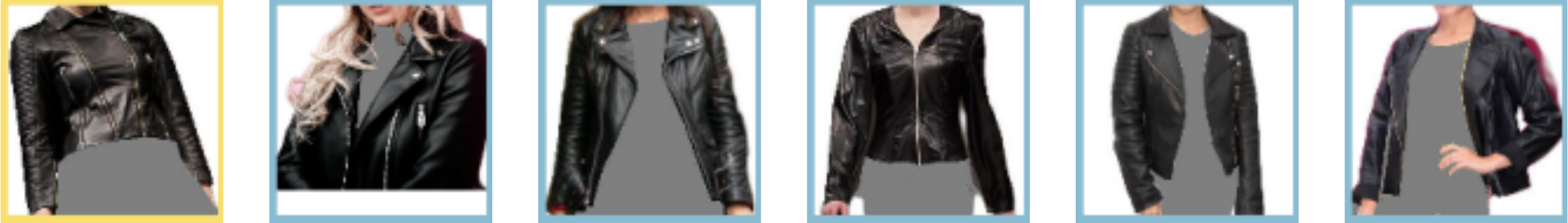}
}
\subfigure[is a \underline{leggings} and has \underline{curved fit silhouette}]{
\includegraphics[width=\linewidth]{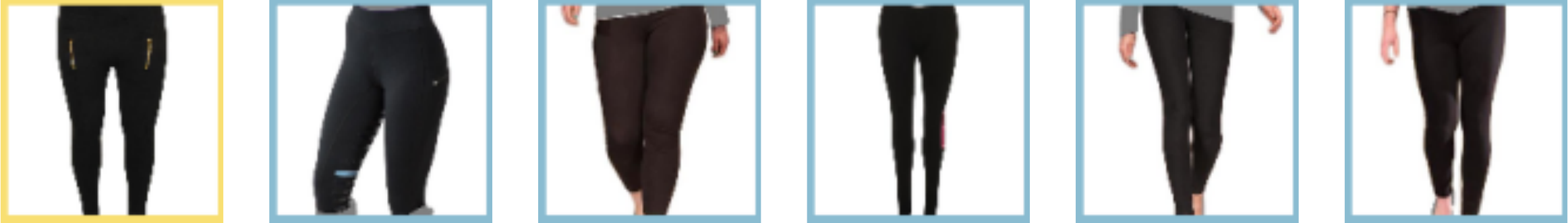}
}
\subfigure[is a \underline{tank top} and is \underline{short length}]{
\includegraphics[width=\linewidth]{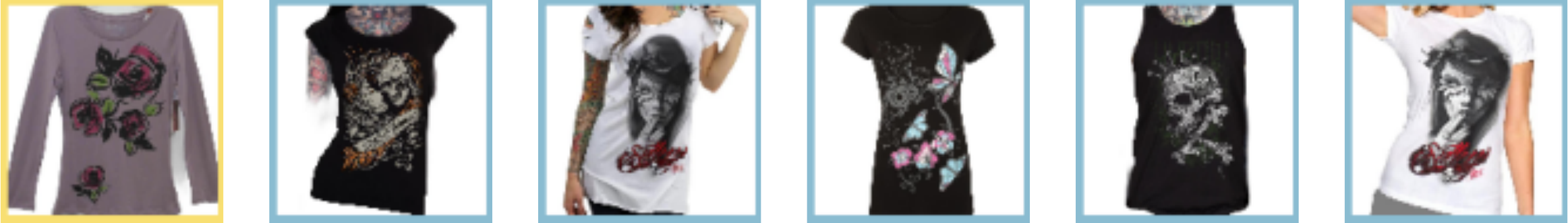}
}
\subfigure[change \underline{non-textile material} to \underline{plastic} and change \underline{decorations} to \underline{ruffle}]{
\includegraphics[width=\linewidth]{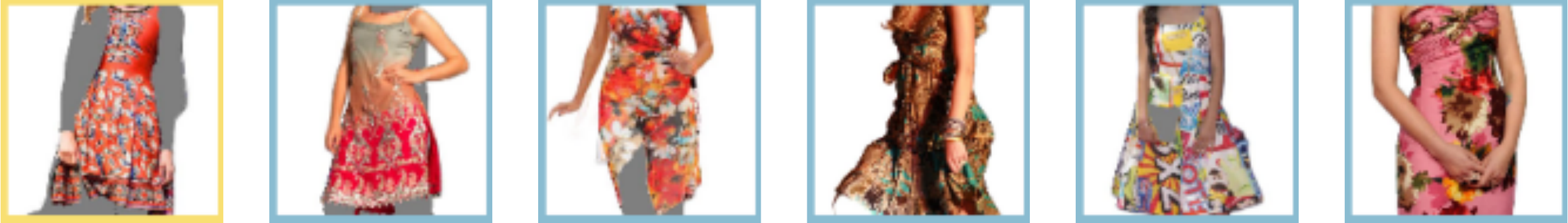}
}
\caption{Failure cases of our multi-task model on TGR subset.}
\label{fig:tgr_failure_cases}
\end{figure}

\begin{figure}[ht]
\centering
\subfigure[replace this \underline{shoe} with a \underline{jacket} that has \underline{lining textile techniques}]{
\includegraphics[width=\linewidth]{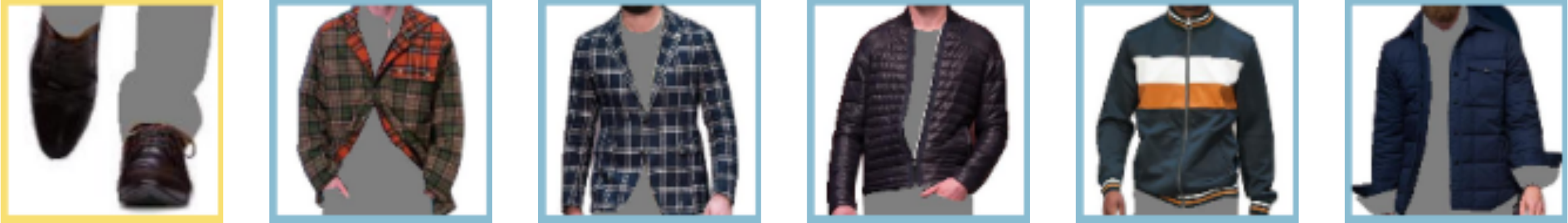}
}
\subfigure[retrieve a \underline{blazer jacket} having a similar style with current \underline{belt}]{
\includegraphics[width=\linewidth]{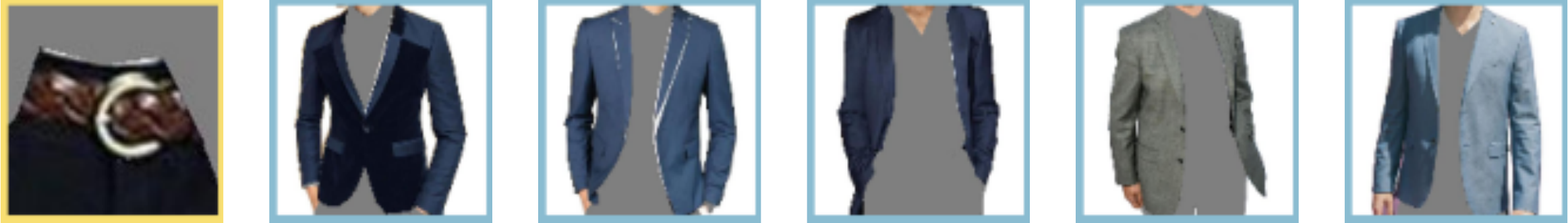}
}
\subfigure[for this \underline{shoe}, find a visually compatible \underline{tights} with \underline{black color}]{
\includegraphics[width=\linewidth]{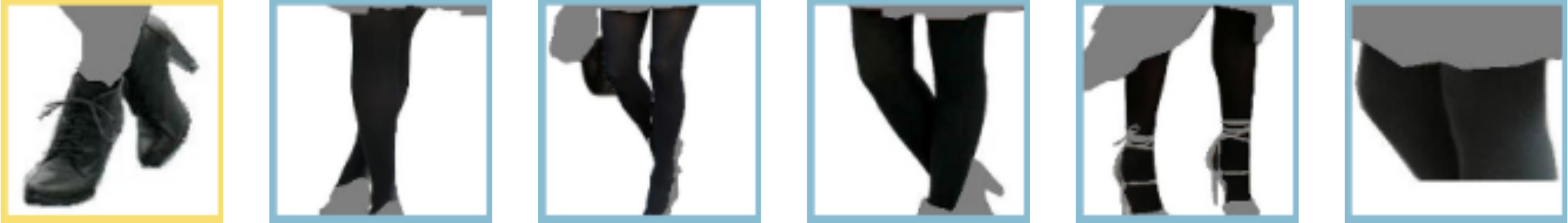}
}
\subfigure[retrieve a \underline{hat} having a similar style with current \underline{bag}]{
\includegraphics[width=\linewidth]{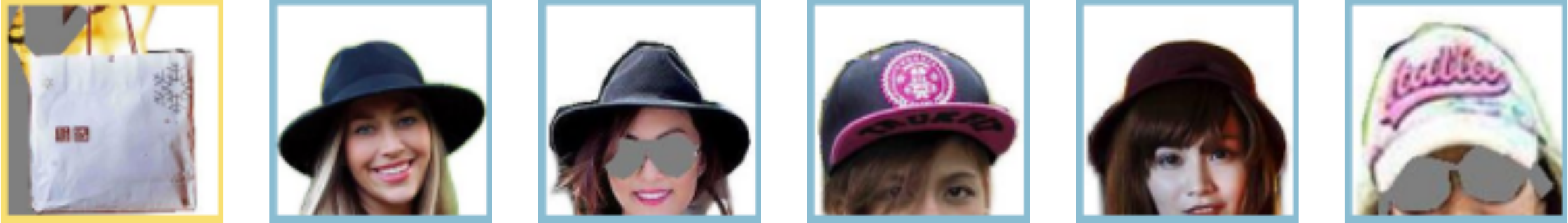}
}
\subfigure[retrieve a \underline{gem dress} having a similar style with current \underline{shoe}]{
\includegraphics[width=\linewidth]{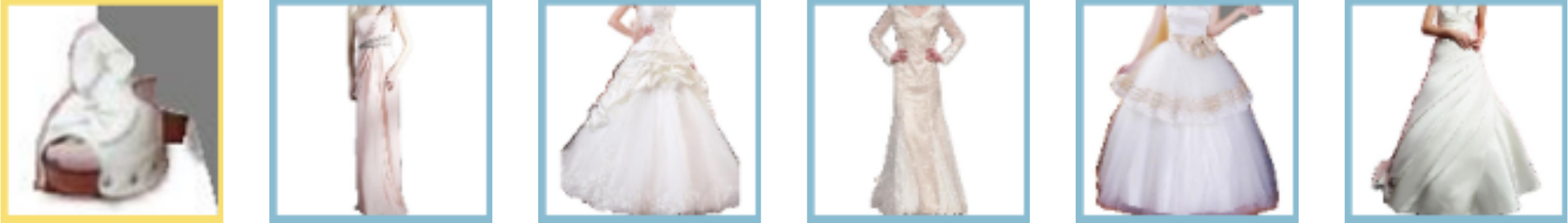}
}
\caption{Failure cases of our multi-task model on VCR subset.}
\label{fig:vcr_failure_cases}
\end{figure}

To better understand the retrieval process of our unified interactive garment retrieval, we visualize some retrieval results in Figure Figure \ref{fig:more_tgr_results} and Figure \ref{fig:more_vcr_results}.
It shows that given a sentence, our model captures both concrete and abstract semantics, including fine-grained attributes and various garment categories.
Besides, many failure cases are also provided in Figure \ref{fig:tgr_failure_cases} and Figure \ref{fig:vcr_failure_cases} to better understand our model's performance.
Even for the failure cases, our model also provides very reasonable predictions.

\subsection{Visualizations of learned latent spaces}
\begin{figure*}[t]
\centering
\subfigure[Visualized latent space of TGR branch.]{
\includegraphics[width=0.8\linewidth]{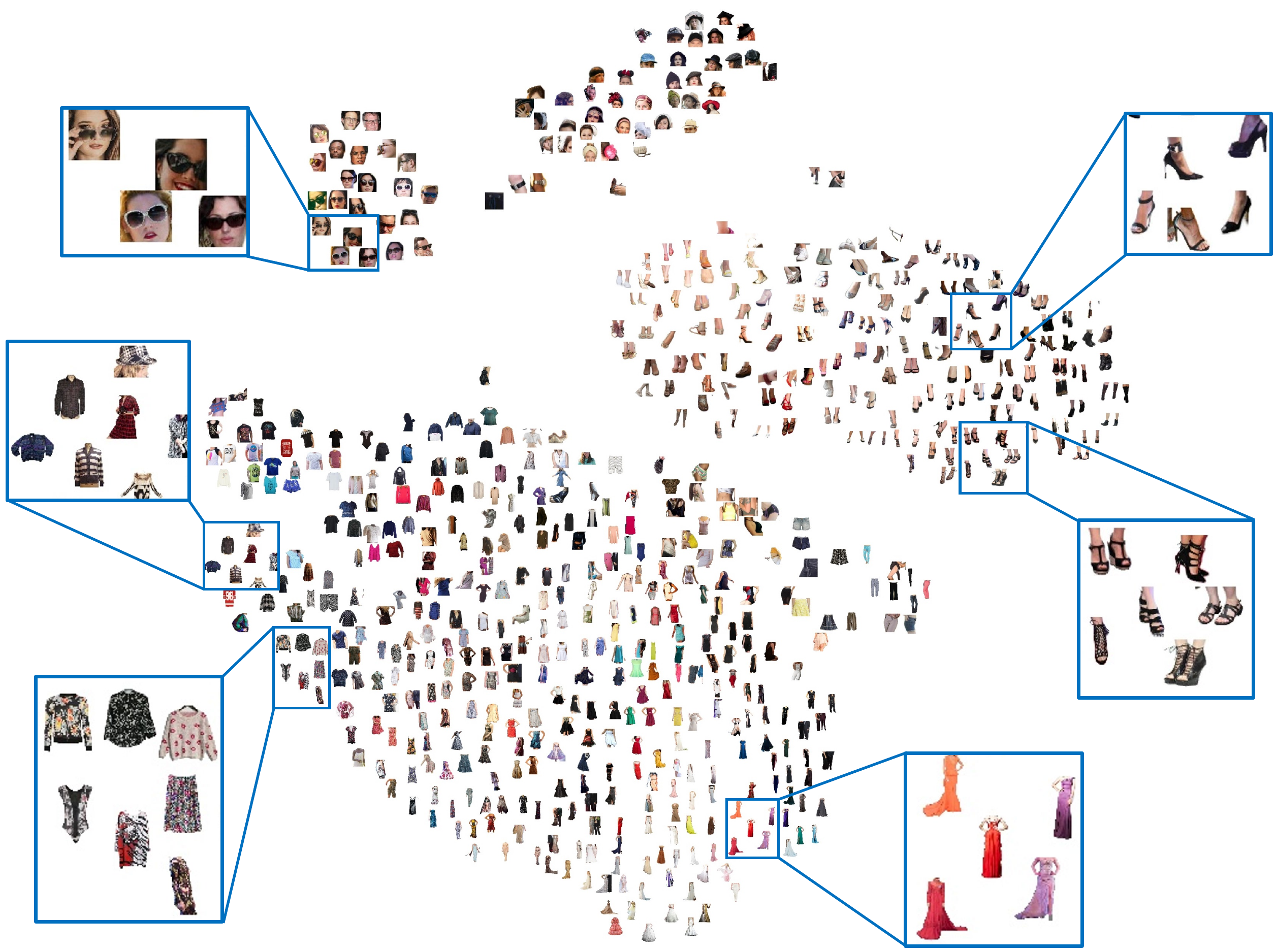}
\label{fig:comp_tsne}
}
\subfigure[Visualized latent space of VCR branch.]{
\includegraphics[width=0.8\linewidth]{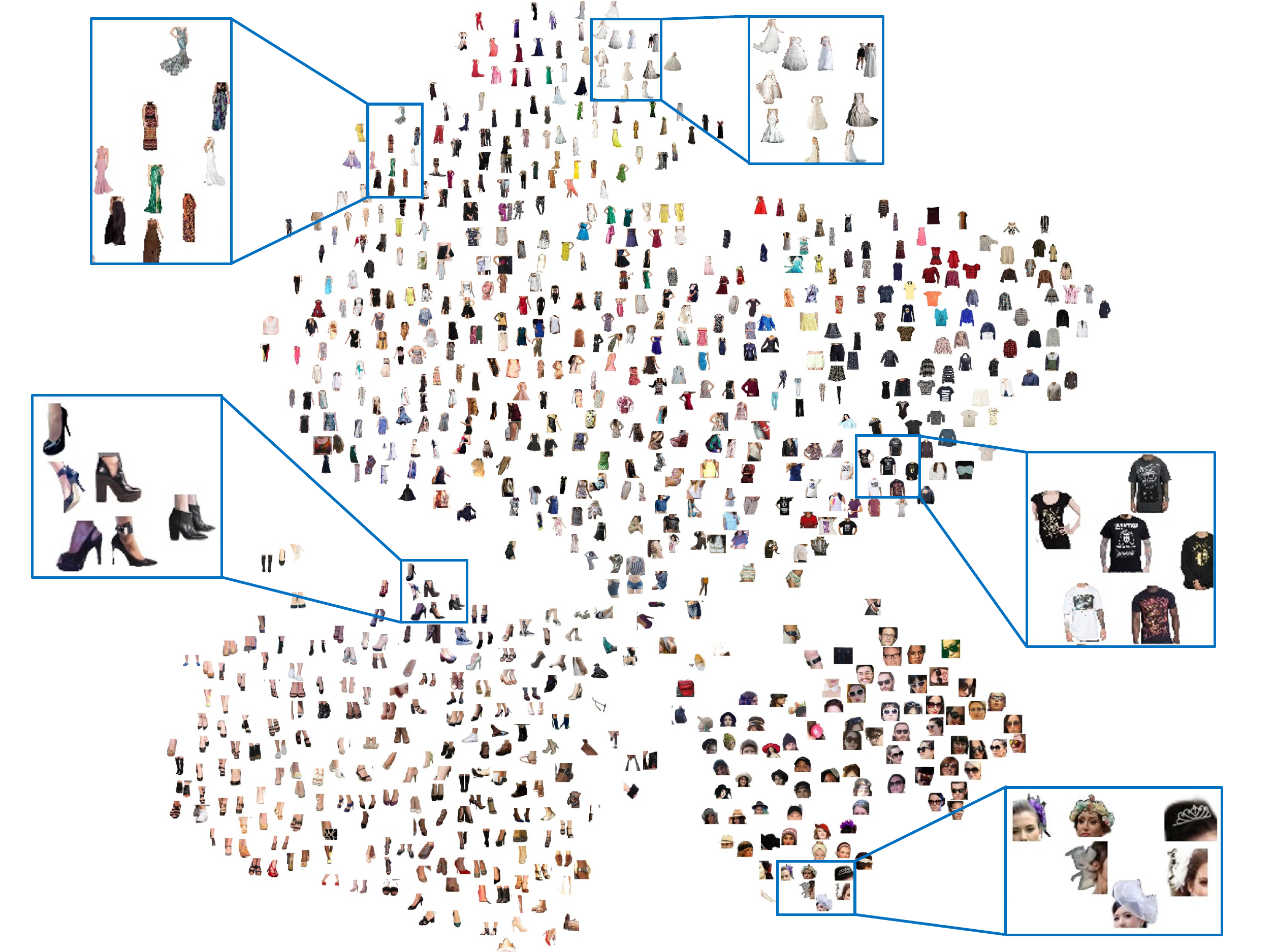}
\label{fig:outfit_tsne}
}

\caption{t-SNE visualizations for two latent spaces learned via our multi-task model.
Best zoom in and view in color.}
\label{fig:tsne}
\end{figure*}

To gain insights into the latent spaces learned by our model, we provide t-SNE \cite{van2008tsne} visualizations for features processed by projection modules in two branches.
Figure \ref{fig:comp_tsne} and \ref{fig:outfit_tsne} illustrate the latent space learned in TGR and VCR branch, respectively.
Both of them demonstrate that our model can learn meaningful latent spaces, where the clusters contain garments with similar appearances.
Specifically, the latent space of the TGR branch mainly focuses on the semantic visual similarity among garments, demonstrating that our TGR branch is superior in learning visual attributes.
Nevertheless, the latent space of the VCR branch does not have a clear boundary as those in the TGR branch.
It seems to pay more attention to the common features of different categories, demonstrating its ability to measure visual compatibility across categories.

\clearpage
\clearpage

\clearpage
\clearpage
{\small
\bibliographystyle{ieee_fullname}
\bibliography{egbib}
}

\end{document}